\documentclass[twoside]{article}

\usepackage[accepted]{aistats2019}

 \newcommand{\citet}{\cite}
\newcommand{\citep}{\cite}

\usepackage{amsmath, amsfonts, bbm, amsthm, mathtools, nicefrac}
\usepackage{color}
\usepackage{appendix}
\usepackage{graphicx, subcaption}
\usepackage{booktabs} %
\usepackage{comment}

\newtheorem{theorem}{Theorem}

\newtheorem{corollary}{Corollary}
\newtheorem{lemma}{Lemma}

\usepackage[utf8]{inputenc}
\usepackage{booktabs} %
\usepackage{enumitem}

\newcommand{\eg}{\textit{e.g.}}
\newcommand{\ie}{\textit{i.e.}}
\usepackage{comment}
\usepackage{tikz,tkz-graph,pgf}
\usetikzlibrary{
  graphs,
  graphs.standard,
  positioning,chains,fit,shapes,calc,
  arrows, automata
}

\begin{document}

\runningtitle{Uncertainty Autoencoders}
\twocolumn[

\aistatstitle{Uncertainty Autoencoders: Learning Compressed Representations via Variational Information Maximization}

\aistatsauthor{ Aditya Grover \And Stefano Ermon }

\aistatsaddress{ Stanford University \And Stanford University } 
]

\begin{abstract}
Compressed sensing techniques enable efficient acquisition and recovery of sparse, high-dimensional data signals via low-dimensional  projections. In this work, we propose \textit{Uncertainty Autoencoders}, a learning framework for unsupervised representation learning inspired by compressed sensing. We treat the low-dimensional projections as noisy latent representations of an autoencoder and directly learn both the acquisition (\ie, encoding) and amortized recovery (\ie, decoding) procedures.
Our learning objective optimizes for a tractable  variational lower bound to the mutual information between the datapoints and the latent representations.
We show how our framework provides a unified treatment to several lines of research in dimensionality reduction, compressed sensing, and generative modeling.
Empirically, we demonstrate a 32\% improvement on average over competing approaches for the task of statistical compressed sensing of high-dimensional datasets.

\end{abstract}
{\let\thefootnote\relax\footnotetext{A preliminary version titled  "Variational Compressive Sensing using Uncertainty Autoencoders" appeared at the Uncertainty in Deep Learning Workshop at UAI, 2018.}}
\section{INTRODUCTION}
The goal of unsupervised representation learning is to learn transformations of the input data which succinctly capture the statistics of an underlying data distribution~\citep{bengio2013representation}.
In this work, we propose a learning framework for unsupervised representation learning inspired by compressed sensing. 
Compressed sensing is a class of techniques used to efficiently acquire and recover high-dimensional data using compressed measurements much fewer than the data dimensionality. 
The celebrated results in compressed sensing posit that \textit{sparse}, high-dimensional datapoints can be acquired using much fewer measurements (roughly logarithmic) than the data dimensionality~\citep{candes2005decoding,donoho2006compressed,candes2006robust}.
The acquisition is done using certain classes of random matrices and the recovery procedure is based on LASSO~\citep{tibshirani1996regression,bickel2009simultaneous}.

The assumptions of sparsity are fairly general and can be applied ``out-of-the-box" for many data modalities, \eg, images and audio are typically sparse in the wavelet and Fourier basis respectively.
However, such assumptions ignore the statistical nature of many real-world problems.
For representation learning in particular, we have access to a training dataset from an underlying domain.
In this work, we use this data to \textit{learn} the acquisition and recovery procedures, thereby sidestepping generic sparsity assumptions.
In particular, we view the compressed measurements 
as the latent representations of an \textit{uncertainty autoencoder}.

An uncertainty autoencoder (UAE) parameterizes both the acquisition and recovery procedures for compressed sensing.
The learning objective for a UAE is based on the InfoMax principle, which seeks to learn encodings that maximize the mutual information between the observed datapoints and noisy representations~\citep{linsker1989generate}. 
Since the mutual information is typically intractable in high-dimensions, we instead maximize tractable variational lower bounds~\citep{barber2003algorithm,alemi2016deep}.
In doing so, we introduce a parameteric decoder that is trained to recover the original datapoint via its noisy representation.
Unlike LASSO-based recovery, a parametric decoder \textit{amortizes} the recovery process, which requires only a forward pass through the decoder  at test time and thus enables scalability to large datasets~\citep{gershman2014amortized,shu2018amortized}.

Notably, the framework of uncertainty autoencoders unifies and extends several lines of prior research in unsupervised representation learning.
First, we show theoretically under suitable assumptions that an uncertainty autoencoder is an \textit{implicit} generative model of the underlying data distribution~\citep{mohamed2016learning}, \ie, a UAE permits sampling from the learned data distribution even though it does not specify an explicit likelihood function.
Hence, it directly contrasts with variational autoencoders (VAE) which specify a likelihood function (which is intractable and approximated by a tractable evidence lower bound)~\citep{kingma-iclr2014}. 
Unlike a VAE, a UAE does not require specifying a prior over the latent representations and hence offsets pathologically observed scenarios that cause the latent representations to be uninformative when used with expressive decoders~\citep{chen2016variational}.

Next, we show that an uncertainty autoencoder, under suitable assumptions, is a generalization of principal component analysis (PCA). While earlier results connecting standard autoencoders with PCA assume linear encodings and decodings~\citep{bourlard1988auto, baldi1989neural,hinton2006reducing}, our result surprisingly holds even for non-linear decodings.
In practice, \emph{linear} encodings learned jointly with non-linear decodings based on the UAE objective vastly outperform the linear encodings obtained via PCA. 
For dimensionality reduction on the MNIST dataset, we observed an average improvement of $5.33\%$ over PCA when the low-dimensional representations are used for classification under a wide range of settings.

We evaluate UAEs for statistical compressed sensing of high-dimensional datasets. 
On the MNIST, Omniglot, and CelebA datasets, we observe average improvements of $38\%$, $31\%$, and $28\%$ in recovery over the closest benchmark across all measurements considered. 
Finally, we demonstrate that uncertainty autoencoders demonstrate good generalization performance across domains in experiments where the encoder/decoder trained on a source dataset are transferred over for compressed sensing of another target dataset.

\section{PRELIMINARIES}
We use upper case to denote probability distributions and assume they admit absolutely continuous densities on a suitable reference measure, denoted by lower case notation. We also use upper and lower case for random variables and their realizations respectively.

\paragraph{Compressed sensing (CS).} Let the datapoint and measurements be denoted with multivariate random variables $X \in \mathbb{R}^n $ and $Y \in \mathbb{R}^m$ respectively.
The goal is to recover $X$ given the measurements $Y$.
For the purpose of compressed sensing, we assume $m < n$ and relate these variables through a measurement matrix $W \in \mathbb{R}^{m \times l}$ and a parameterized acquisition function $f_\psi: \mathbb{R}^n \rightarrow \mathbb{R}^l$ (for any integer $l>0$) such that:
\begin{align}\label{eq:non_linear_system}
y &= W f_\psi(x) + \epsilon
\end{align}
where $\epsilon$ is the measurement noise.
If we let $f_\psi(\cdot)$ be the identity function (\textit{i.e.}, $f_\psi(x) = x$ for all $x$), then we recover a standard system of underdetermined linear equations where measurements are linear combinations of the datapoint corrupted by noise. 
In all other cases, the acquisition function transforms $x$ such that $f_\psi(x)$ is potentially more amenable for compressed sensing. 
For instance, $f_\psi(\cdot)$ could specify a change of basis that encourages sparsity, \textit{e.g.}, a Fourier basis for audio. 
Note that we allow the codomain of the mapping $f_\psi(\cdot)$ to be defined on a higher or lower dimensional space (\textit{i.e.}, $l \neq n$ in general).

\paragraph{Sparse CS.}
To obtain nontrivial solutions to an underdetermined system, $X$ is assumed to be sparse in some basis $B$.
We are not given any additional information about $X$. 
The measurement matrix $W$ is a random Gaussian matrix and the recovery is done via LASSO~\citep{candes2005decoding,donoho2006compressed,candes2006robust}.
LASSO solves for a convex $\ell_1$-minimization  
problem such that the reconstruction $\widehat{x}$ for any datapoint $x$ is given as:
$
\hat{x}=\mathrm{arg} \min_{x} \Vert Bx \Vert_1
+ \lambda \Vert y- Wx \Vert_2^2$
where $\lambda>0$ is a tunable hyperparameter. 

\paragraph{Statistical CS.} 
In \textit{statistical} compressed sensing~\citep{yu2011statistical}, we are additionally given access to a set of signals $\mathcal{D}$,
 such that each $x\in \mathcal{D}$ is assumed to be sampled i.i.d. from a data distribution $Q_\mathrm{data}$. 
 Using this dataset, we learn the the measurement matrix $W$ and the acquisition function $f_\psi(\cdot)$ in Eq.~\eqref{eq:non_linear_system}. 

At test time, we directly observe the measurements $y_{\mathrm{test}}$ that are assumed to satisfy Eq.~\eqref{eq:non_linear_system} for a target datapoint $x_{\mathrm{test}}\sim Q_\mathrm{data}(X)$ and the task is to provide an accurate reconstruction $\hat{x}_{\mathrm{test}}$. Evaluation is based on the reconstruction error between $x_{\mathrm{test}}$ and $\hat{x}_{\mathrm{test}}$. 
Particularly relevant to this work, we can optionally \textit{learn} a recovery function $g_\theta: \mathbb{R}^m \rightarrow \mathbb{R}^n$ to reconstruct $X$ given the measurements $Y$. 

This amortized approach~\cite{shu2018amortized} is in contrast to standard LASSO-based decoding which solves an optimization problem \emph{for every new datapoint} at test time.
If we learned the recovery function $g_\theta(\cdot)$ during training, then $\hat{x}_{\mathrm{test}}=g_\theta(y_{\mathrm{test}})$ and the $\ell_2$ error is given by $\Vert x_{\mathrm{test}} -  g_\theta(y_{\mathrm{test}})\Vert_2$. 
Such a recovery process requires only a function evaluation at test time and permits scaling to large datasets~\citep{gershman2014amortized,shu2018amortized}.

\paragraph{Autoencoders.} An autoencoder is a pair of parameterized functions $(e,d)$ designed to encode and decode datapoints. 
For a standard autoencoder, let $e: \mathbb{R}^n \rightarrow \mathbb{R}^m$ and $d: \mathbb{R}^m \rightarrow \mathbb{R}^n$ denote the encoding and decoding functions respectively for an $n$-dimensional datapoint and an $m$-dimensional latent space. 
The learning objective minimizes the $l_2$ reconstruction error over a dataset $\mathcal{D}$:
\begin{align}
\min_{e,d} \sum_{x \in \mathcal{D}} \Vert x-d(e(x))\Vert_2^2
\end{align}
where the encoding and decoding functions are 
typically parameterized using neural networks.

\section{UNCERTAINTY AUTOENCODER}\label{sec:framework}

Consider a joint distribution between the signals $X$ and the measurements $Y$, which factorizes as $Q_\phi(X,Y) = Q_{\mathrm{data}}(X)Q_\phi(Y\vert X)$. Here, $Q_{\mathrm{data}}(X)$ is a fixed data distribution and $Q_\phi(Y\vert X)$ is a parameterized observation model that depends on the measurement noise $\epsilon$, as given by Eq.~\eqref{eq:non_linear_system}. In particular, $\phi$ corresponds to collectively the set of measurement matrix parameters $W$ and the acquisition function parameters $\psi$.  For instance, for isotropic Gaussian noise $\epsilon$ with a fixed variance $\sigma^2$, we have $Q_\phi(Y\vert X) = \mathcal{N}(Wf_\psi(X), \sigma^2 I_m)$.

In an uncertainty autoencoder, we wish to learn the parameters $\phi$ that permit efficient and accurate recovery of a signal $X$ using the measurements $Y$. In order to do so, we propose to maximize the mutual information between $X$ and $Y$:
\begin{align}\label{eq:mi_2}
\max_\phi &\;\; I_\phi(X,Y) = \int q_\phi(x, y) \log \frac{q_\phi(x, y)}{q_\mathrm{data}(x) q_\phi(y)}\mathrm{d}x \mathrm{d}y\nonumber\\
&= H(X) - H_\phi(X \vert Y)
\end{align}
where $H$ denotes differential entropy.
The intuition is simple: if the measurements preserve 
maximum information about the signal, we can hope that recovery will have low reconstruction error. We formalize this intuition by noting that this objective is equivalent to maximizing the average log-posterior probability of $X$ given $Y$.
In fact, in Eq.~\eqref{eq:mi_2}, we can omit the term corresponding to the data entropy (since it is independent of $\phi$) to get the following equivalent objective:
\begin{align}\label{eq:cond_entropy}
\max_\phi - H_\phi(X \vert Y) = \mathbb{E}_{Q_\phi(X,Y)}[\log q_\phi(x \vert y)].
\end{align}

Even though the mutual information is maximized and equals the data entropy when $Y=X$, the dimensionality constraints on $m \ll n$, the parametric assumptions on $f_\psi(\cdot)$,
and the noise model prohibit learning an identity mapping. 
Note that the properties of noise $\epsilon$ such as the distributional family and sufficient statistics are externally specified.
For example, these could be specified based on properties of the measurement device for compressed sensing.
More generally for unsupervised representation learning, we treat these properties as hyperparameters tuned based on the reconstruction loss on a held-out set, or any other form of available supervision.
It is not suggested to optimize for these statistics during learning as the UAE would tend to shrink this noise to zero to maximize mutual information, thus ignoring measurement uncertainty in the context of compressed sensing and preventing generalization to out-of-distribution examples for representation learning.
The theoretical results in Section~\ref{sec:analysis} analyze the effect of noise more formally.

Estimating mutual information between arbitrary high dimensional random variables can be challenging. 
However, we can lower bound the mutual information by introducing a variational approximation to the model posterior $Q_\phi(X \vert Y)$~\citep{barber2003algorithm}. Denoting this approximation as $P_\theta(X \vert Y)$, we get the following lower bound:
\begin{align}\label{eq:mi_lower_bound}
I_\phi(X,Y)  &\geq H(X) + \mathbb{E}_{Q_\phi(X,Y)}\left[\log p_\theta(x\vert y)\right].
\end{align}
Comparing Eqs.~(\ref{eq:mi_2}, \ref{eq:cond_entropy}, \ref{eq:mi_lower_bound}), we can see that the second term in Eq.~\eqref{eq:mi_lower_bound} approximates the intractable negative conditional entropy, $-H_\phi(X \vert Y)$ with a variational lower bound.
Optimizing this bound leads to a decoding distribution given by $P_\theta(X\vert Y)$ with variational parameters $\theta$.
The bound is tight when there is no distortion during recovery, or equivalently when the decoding distribution $P_\theta(X\vert Y)$ matches the true posterior $Q_\phi(X \vert Y)$ (\textit{i.e.}, the Bayes optimal decoder).

\paragraph{Stochastic optimization.}
Formally, the uncertainty autoencoder (UAE) objective is given by:
\begin{align}\label{eq:uae_obj}
\max_{\theta, \phi}\mathbb{E}_{Q_\phi(X,Y)}\left[\log p_\theta(x\vert y)\right]. 
\end{align}
In practice, the data distribution $Q_{\mathrm{data}}(X)$ is unknown and accessible only via a finite dataset $\mathcal{D}$. Hence, expectations with respect to $Q_{\mathrm{data}}(X)$ and its gradients can be estimated
using Monte Carlo methods. This allows us to express the UAE objective as:
\begin{align}\label{eq:uae_obj_emp}
\max_{\theta,\phi} \;\;&\sum_{x \in \mathcal{D}}\mathbb{E}_{Q_\phi(Y\vert x)}\left[\log p_\theta(x\vert y) 
\right]
:=\mathcal{L}(\phi, \theta
; 
\mathcal{D}).
\end{align}

Tractable evaluation of the above objective is closely tied to the distributional assumptions on the noise model. 
This could be specified externally based on, \eg, properties of the sensing device in compressed sensing.
For the typical case of an isotropic Gaussian noise model, we know that $Q_\phi(Y\vert X) = \mathcal{N}(Wf_\psi(X), \sigma^2 I_m)$, which is easy-to-sample. 

While Monte Carlo gradient estimates with respect to $\theta$ can be efficiently obtained via linearity of expectation,
gradient estimation with respect to $\phi$ is challenging since these parameters specify the sampling distribution $Q_\phi(Y \vert X)$. 
One solution is to evaluate \textit{score function} gradient estimates along with control variates~\citep{fu2006gradient,glynn1990likelihood,williams1992simple}. Alternatively, many continuous distributions (\textit{e.g.}, the isotropic Gaussian and Laplace distributions) can be \textit{reparameterized} such that it is possible to obtain samples by applying a deterministic transformation to samples from a fixed distribution and typically leads to low-variance gradient estimates~\citep{kingma-iclr2014,rezende2014stochastic,glasserman2013monte,schulman2015gradient}. 

\section{THEORETICAL ANALYSIS}\label{sec:analysis}

In this section, we derive connections of uncertainty autoencoders with generative modeling and Principal Component Analysis (PCA). The proofs of all theoretical results in this section are in Appendix~\ref{app:proof}.
\subsection{Implicit generative modeling}
Starting from an arbitrary point $x^{(0)} \in \mathbb{R}^n$, define a Markov chain over $X,Y$ with the following transitions:
\begin{align}
y^{(t)} &\sim Q_\phi(Y \vert x^{(t)})\label{eq:markov_chain_1}\\
x^{(t+1)} &\sim P_\theta(X \vert y^{(t)})\label{eq:markov_chain_2}
\end{align}
\begin{theorem}\label{thm:markov_chain}
Let $\theta^\ast,\phi^\ast$ denote an optimal solution to the UAE objective in Eq.~\eqref{eq:uae_obj}.
If there exists a $\phi$ such that $q_\phi(x|y)=p_{\theta^\ast}(x|y)$
and the Markov chain defined in Eqs.~(\ref{eq:markov_chain_1}, \ref{eq:markov_chain_2}) is ergodic, then the stationary distribution of the chain for the parameters $\phi^\ast$ and $\theta^\ast$ is given by $Q_{\phi^\ast}(X,Y)$.
\end{theorem}

The above theorem suggests an interesting insight into the behavior of UAEs. Under idealized conditions, the learned model specifies an implicit generative model for $Q_{\phi^\ast}(X,Y)$. Further, ergodicity can be shown to hold for the isotropic Gaussian noise model.

\begin{corollary}\label{thm:markov_chain_gaussian}
Let $\theta^\ast,\phi^\ast$ denote an optimal solution to the UAE objective in Eq.~\eqref{eq:uae_obj}.
If there exists a $\phi$ such that $q_\phi(x|y)=p_{\theta^\ast}(x|y)$ and the noise model is Gaussian, then
the stationary distribution of the chain for the parameters $\phi^\ast$ and $\theta^\ast$ is given by $Q_{\phi^\ast}(X,Y)$.
\end{corollary}

The marginal of the joint distribution $Q_{\phi}(X,Y)$ with respect to $X$ corresponds to the data distribution. 
A UAE hence seeks to learn an \textit{implicit} generative model of the data distribution~\citep{diggle1984monte,mohamed2016learning}, \textit{i.e.}, even though we do not have a tractable estimate for the likelihood of the model, we can generate samples using the Markov chain transitions defined in Eqs.~(\ref{eq:markov_chain_1}, \ref{eq:markov_chain_2}). 

\subsection{Optimal encodings}

A UAE can also be viewed as a dimensionality reduction technique for the dataset $\mathcal{D}$.
While in general the encoding performing this reduction can be nonlinear, the case of a linear encoding 
is one 
where the projection vectors are given as the rows of the measurement matrix $W$. 
The result below characterizes the optimal encoding of the dataset $\mathcal{D}$ with respect to the UAE objective for an isotropic Gaussian noise model.

\begin{theorem}\label{thm:pca}
Assume a uniform data distribution over a finite dataset $\mathcal{D}$. Further, we assume that expectations in the UAE objective exist, and the signals and measurement matrices are bounded in $\ell_2$/Frobenius norms, \textit{i.e.,} $\Vert x\Vert_2 \leq k_1$ for all $x \in \mathcal{D}$, $\Vert W\Vert_F \leq k_2$ for some positive constants $k_1, k_2 \in \mathbb{R}^{+}$.
For a linear encoder 
and isotropic Gaussian noise $\epsilon\sim \mathcal{N}(0, \sigma^2 I)$, the optimal measurement matrix $W^\ast$ 
that maximizes the mutual information for an optimal decoder 
in the limit $\sigma \rightarrow \infty$ 
is given as:
\begin{align*}
W^\ast = 
\mathrm{eig}_m \left( 
\sum_{x_i,x_j\in \mathcal{D}} 
\left[ (x_i-x_j)(x_i-x_j)^T\right] \right)
\end{align*}
where $\mathrm{eig}_m(M)$ denotes the top-$m$ eigenvectors of the matrix $M$ with the largest eigenvalues (specified up to a positive scaling constant).
\end{theorem}

\begin{figure}[t]
\centering
\includegraphics[width=0.75\columnwidth]{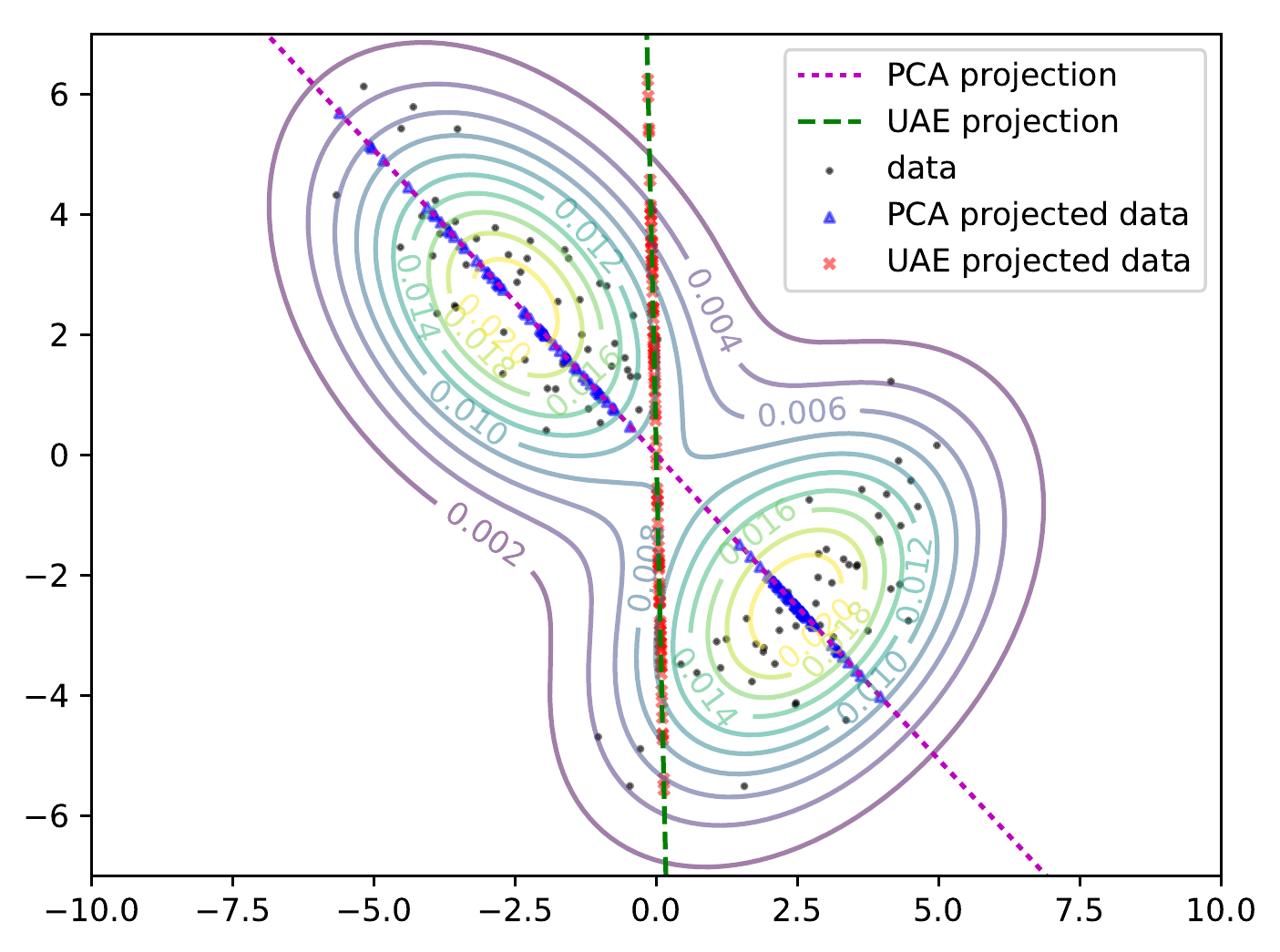}
\caption{Dimensionality reduction using PCA vs. UAE. Projections of the data (\textbf{black points}) on the UAE direction (\textbf{green line}) maximize the likelihood of decoding unlike the PCA projection axis (\textbf{magenta line}) which collapses many points in a narrow region.}\label{fig:gmm}
\vspace{-0.1in}
\end{figure}

\begin{figure*}[th]
\centering
\begin{subfigure}[b]{0.35\textwidth}
\centering
\includegraphics[width=\textwidth]{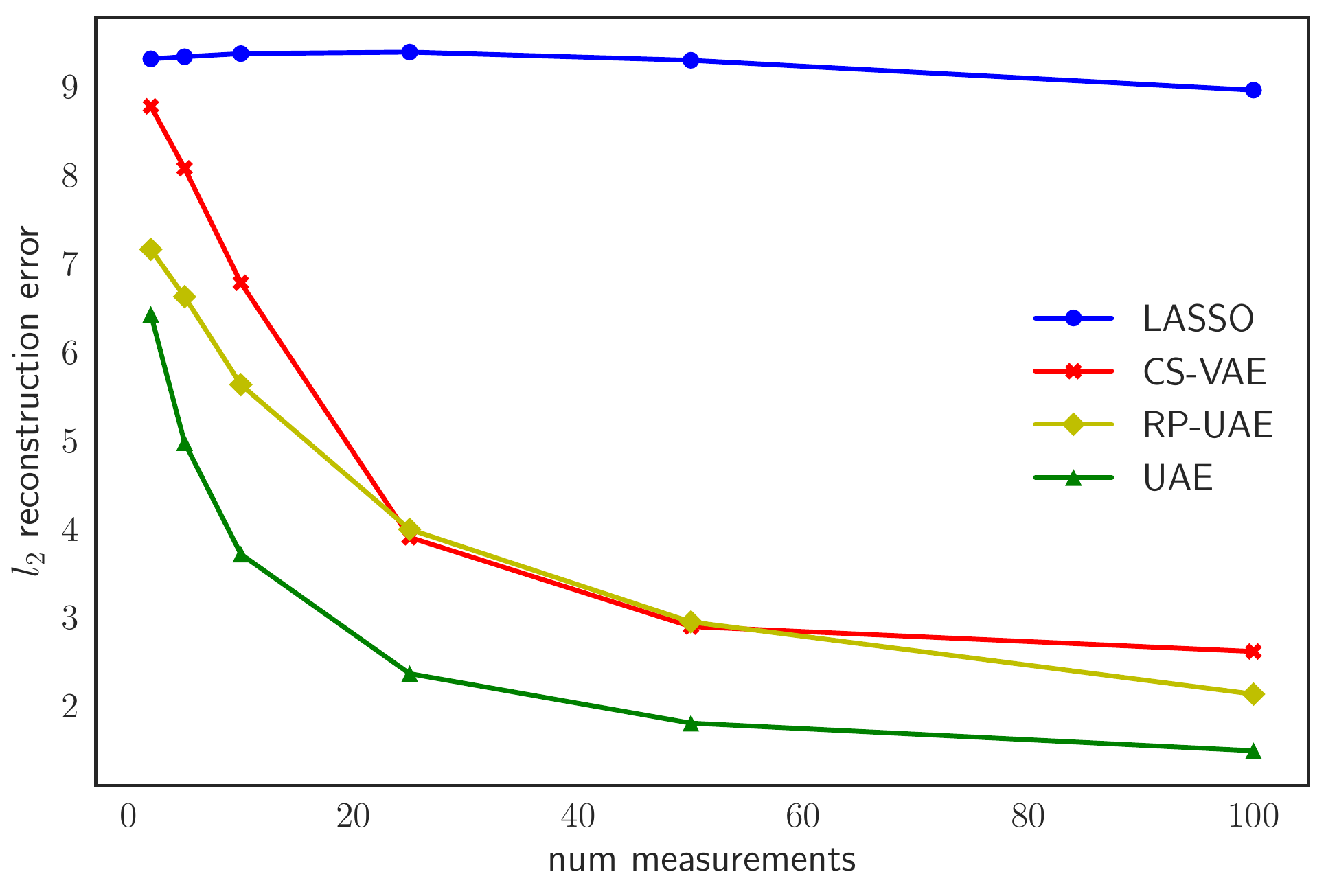}
\caption{MNIST}
\end{subfigure}
\begin{subfigure}[b]{0.35\textwidth}
\centering
\includegraphics[width=\textwidth]{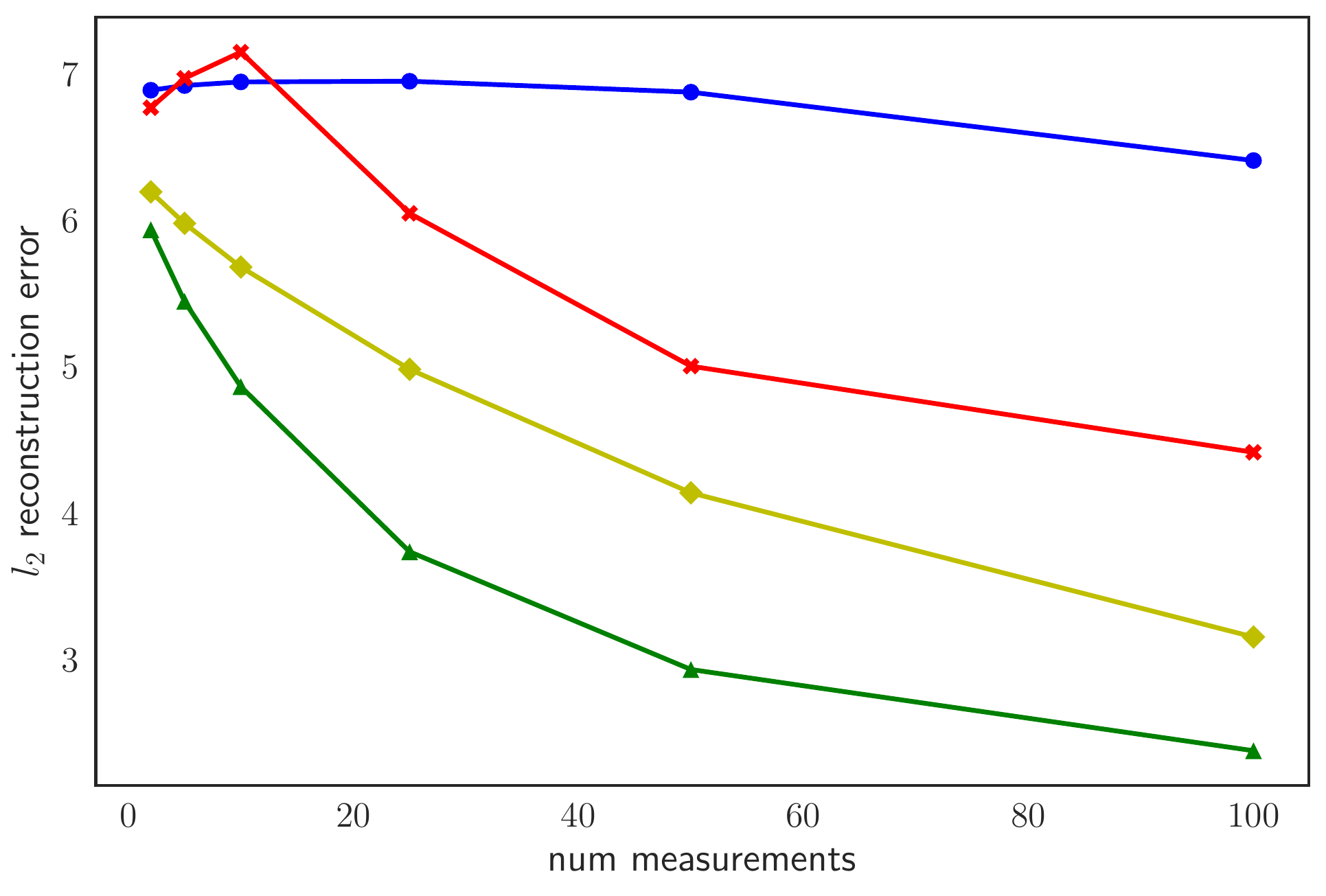}
\caption{Omniglot}
\end{subfigure}
\caption{Test $\ell_2$ reconstruction error (per image) for compressed sensing.}\label{fig:l2_reconstr}
\vspace{-0.1in}
\end{figure*}

\begin{figure*}[t]
\centering
\begin{subfigure}[b]{0.495\textwidth}
\centering
\includegraphics[width=\textwidth]{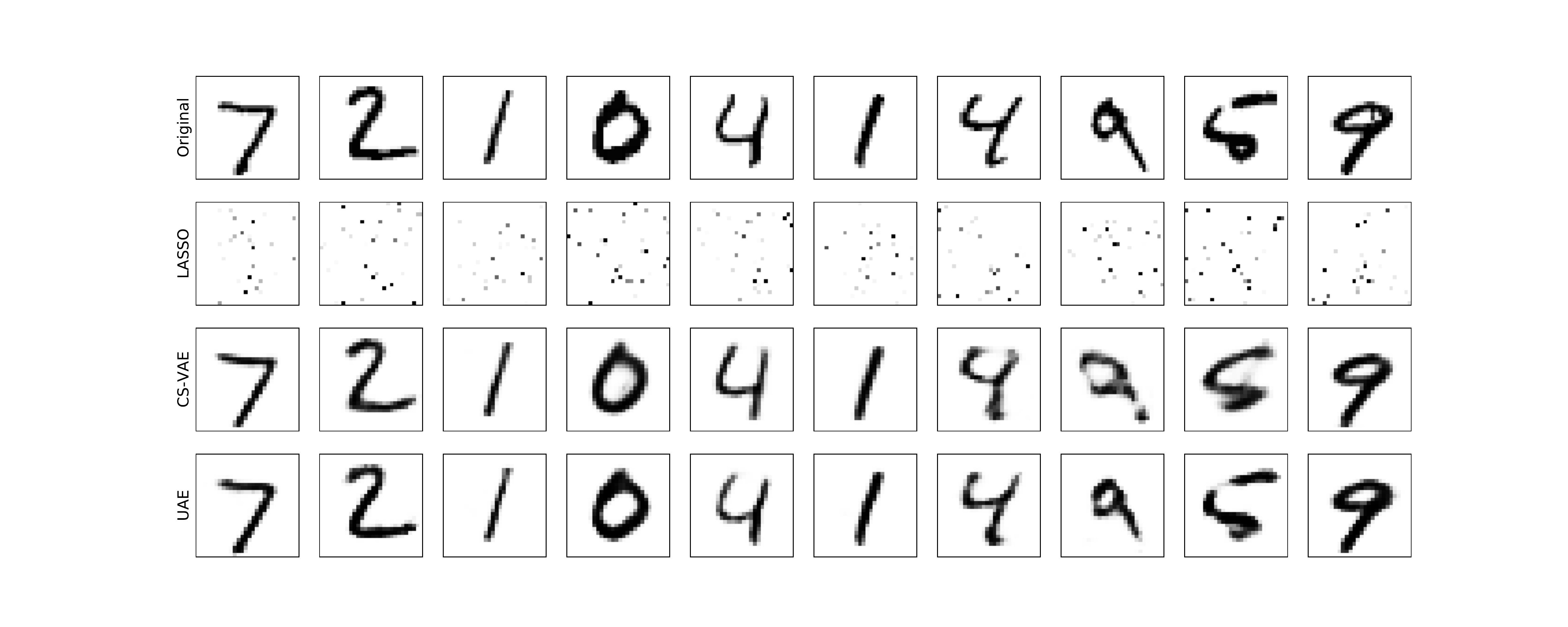}
\caption{MNIST}
\end{subfigure}
\begin{subfigure}[b]{0.495\textwidth}
\centering
\includegraphics[width=\textwidth]{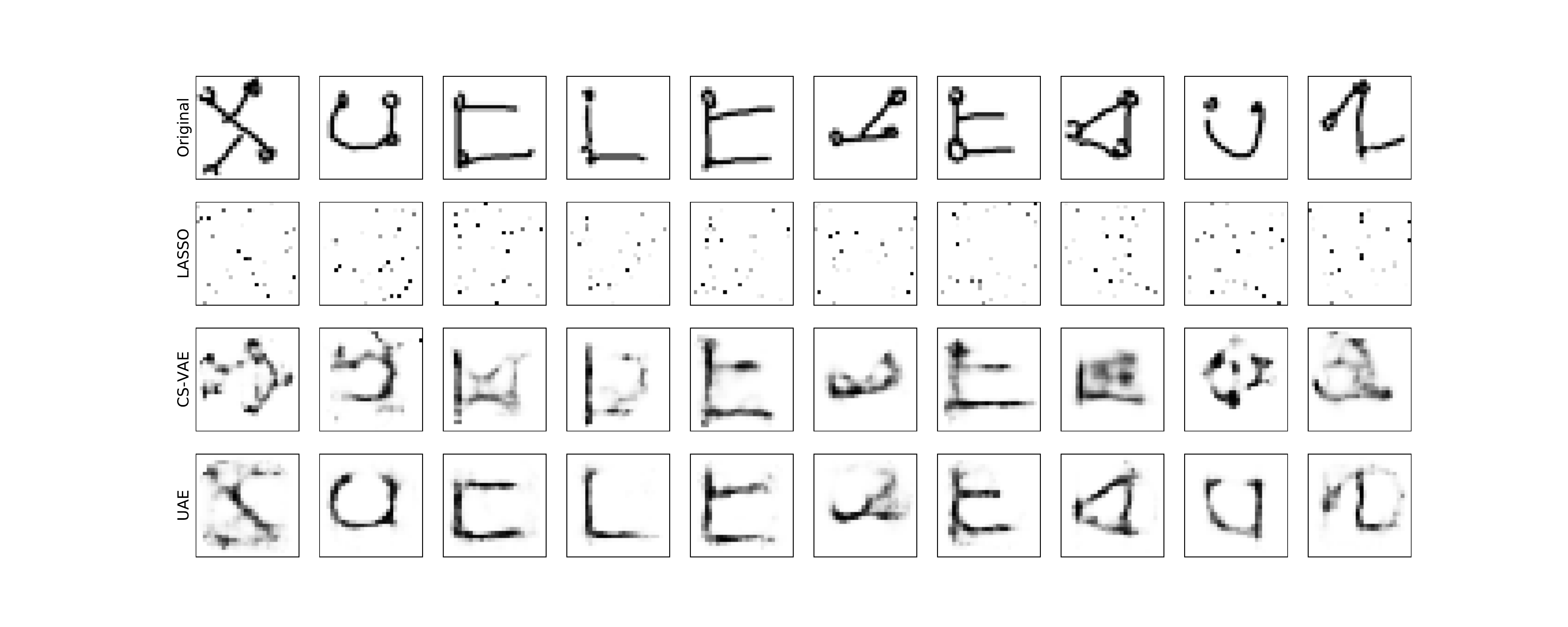}
\caption{Omniglot}
\end{subfigure}
\caption{Reconstructions for $m=25$. \textbf{Top:}  Original. \textbf{Second:} LASSO. \textbf{Third:} CS-VAE. \textbf{Last:} UAE. $25$ projections of the data are sufficient for UAE to reconstruct the original image with high accuracy.}\label{fig:reconstr}
\end{figure*}

\begin{figure*}[t]
\centering
\begin{subfigure}[b]{0.35\textwidth}
\centering
\includegraphics[width=\textwidth]{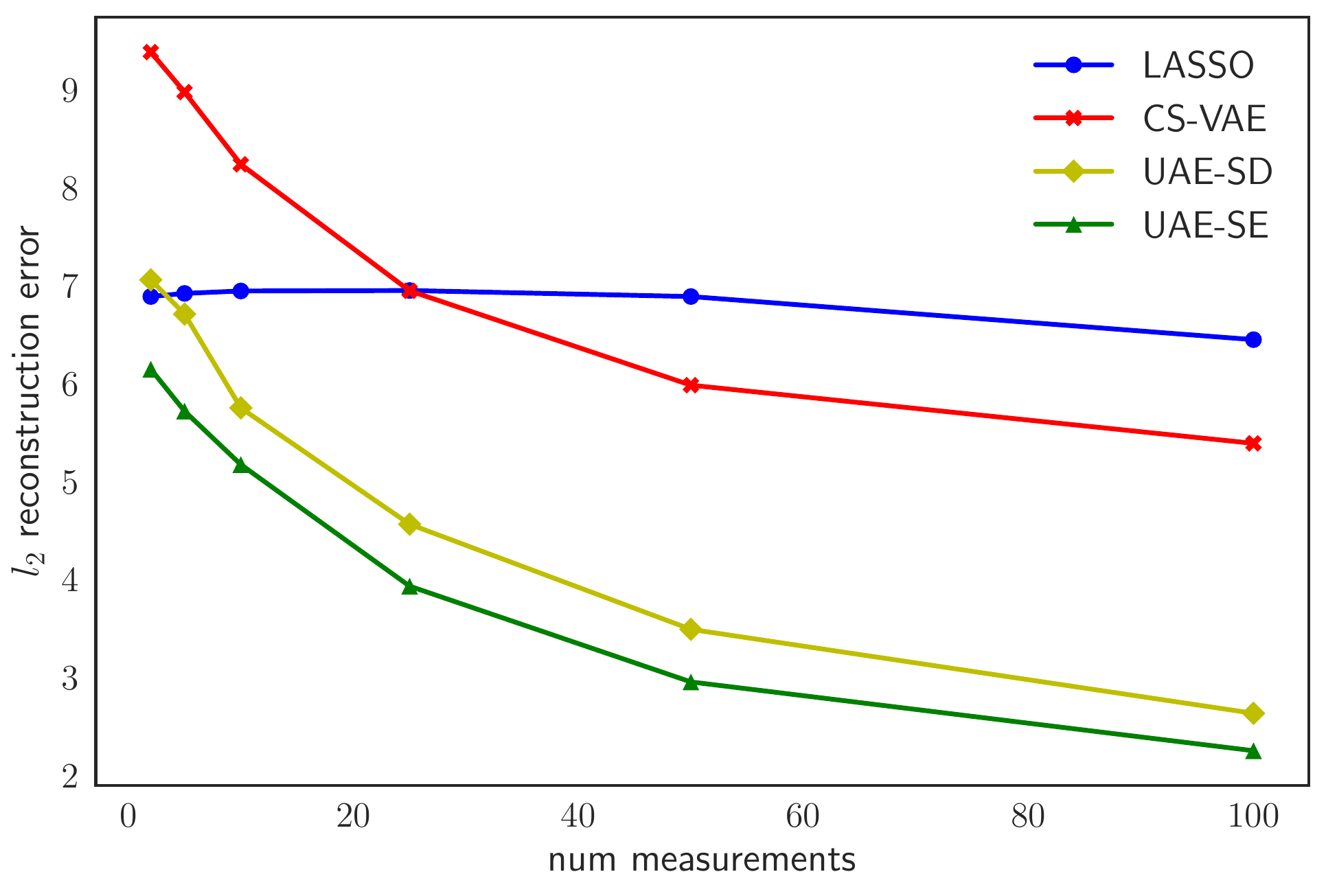}
\caption{Source: MNIST, Target: Omniglot}
\end{subfigure}
\begin{subfigure}[b]{0.35\textwidth}
\centering
\includegraphics[width=\textwidth]{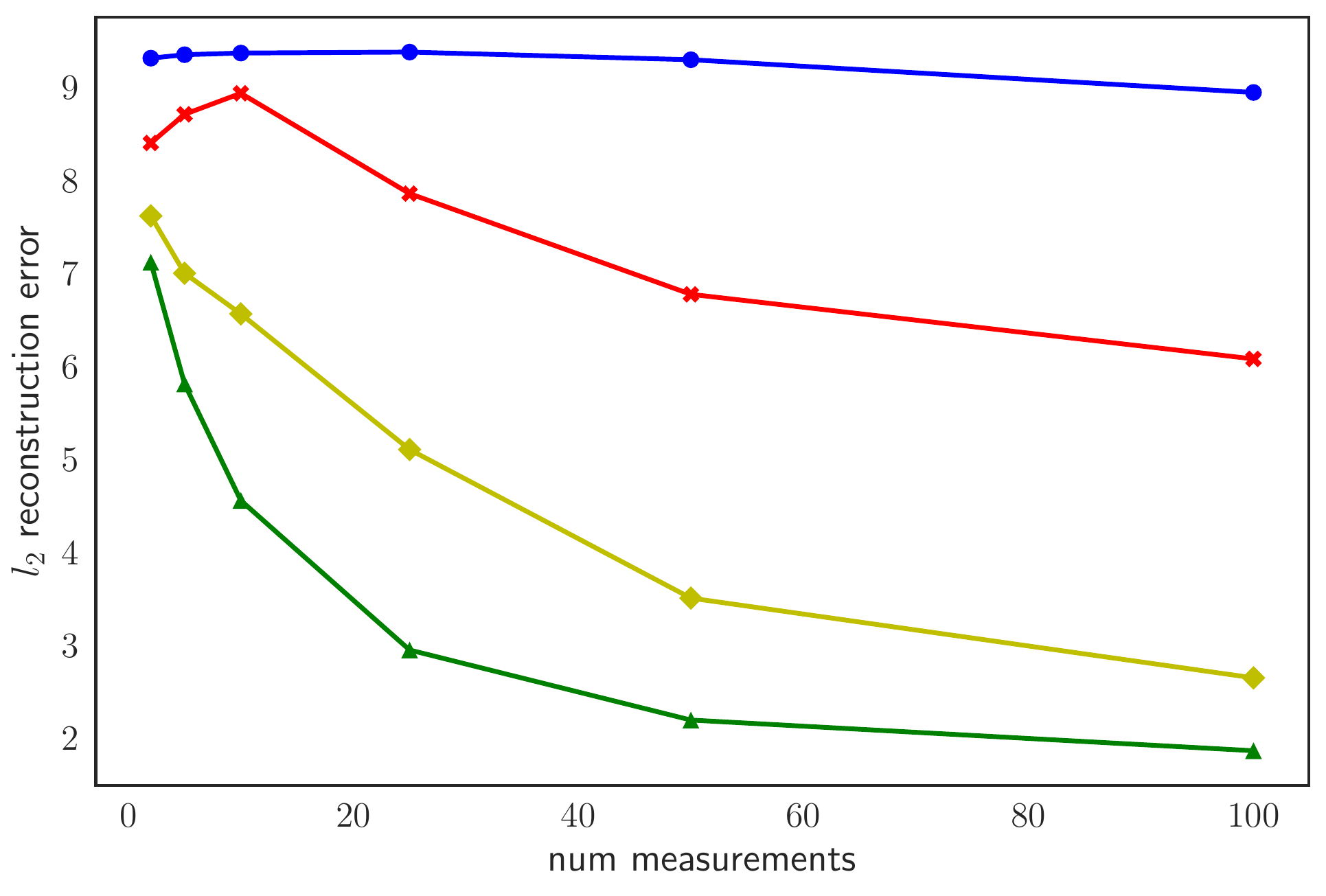}
\caption{Source: Omniglot, Target: MNIST}
\end{subfigure}
\caption{Test $\ell_2$ reconstruction error (per image) for transfer compressed sensing.}\label{fig:l2_reconstr_transfer}
\vspace{-0.1in}
\end{figure*}

\begin{figure*}[t]
\centering
\begin{subfigure}[b]{0.495\textwidth}
\centering
\includegraphics[width=\textwidth]{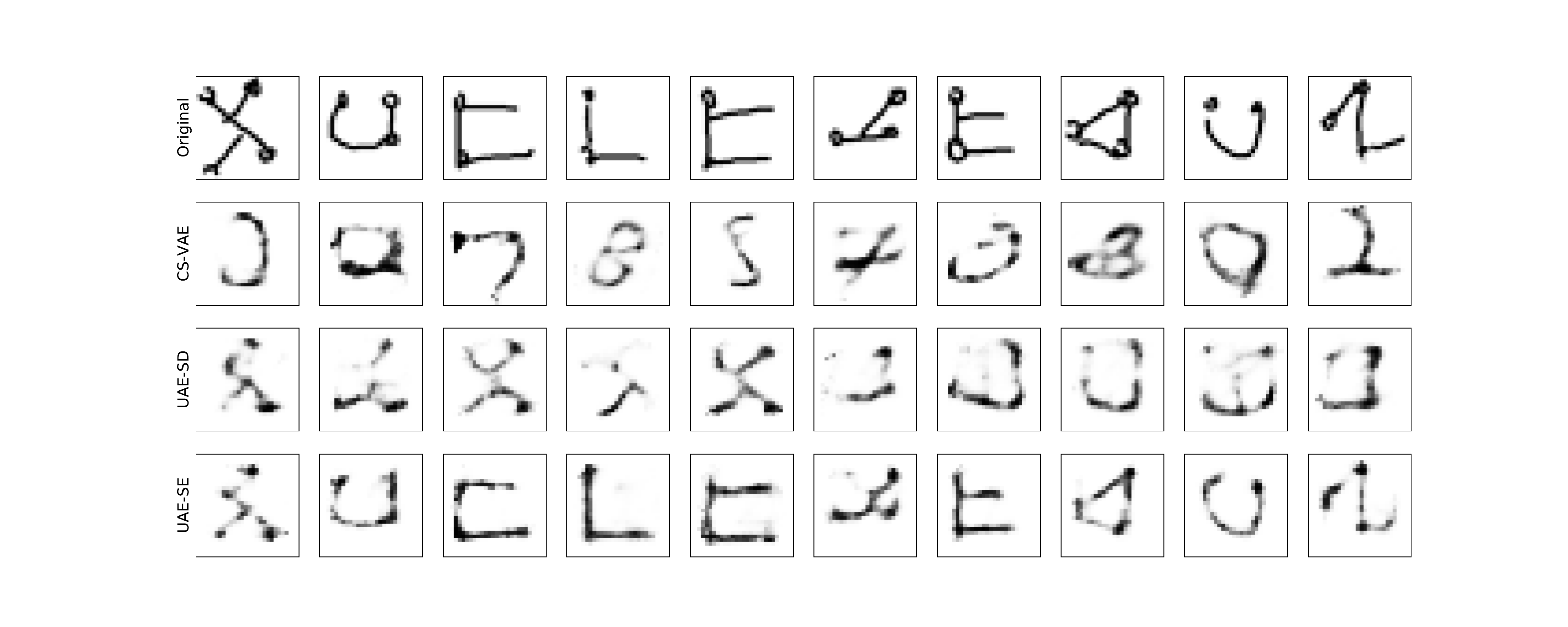}
\caption{Source: MNIST, Target: Omniglot}
\end{subfigure}
\begin{subfigure}[b]{0.495\textwidth}
\centering
\includegraphics[width=\textwidth]{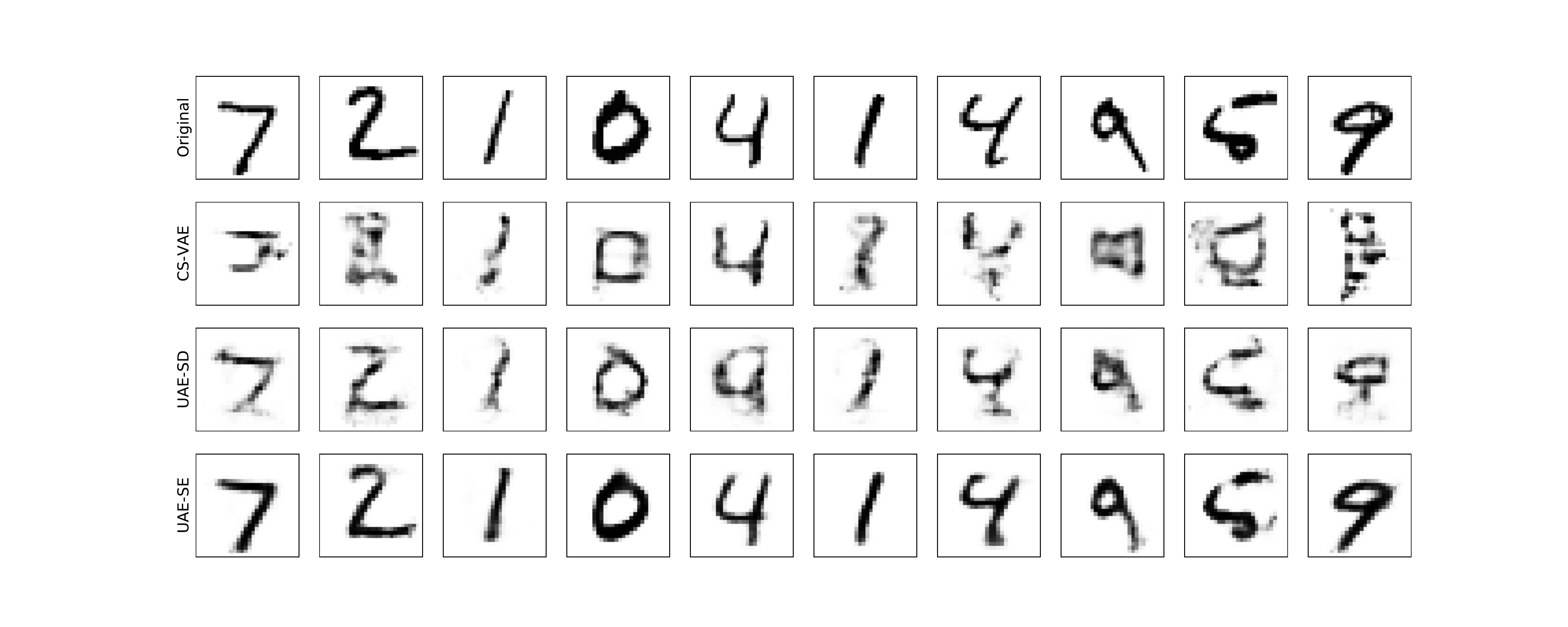}
\caption{Source: Omniglot, Target: MNIST}
\end{subfigure}
\caption{Reconstructions for $m=25$. \textbf{Top:} Target. \textbf{Second:} CS-VAE. \textbf{Third:} UAE-SD. \textbf{Last:} UAE-SE. }
\label{fig:reconstr_transfer}
\end{figure*}

Under the stated assumptions, the above result suggests an interesting connection between UAE and PCA.
PCA seeks to find the directions that explain the most variance in the data. Theorem~\ref{thm:pca} suggests that when the noise in the projected signal is very high, the optimal projection directions (\textit{i.e.}, the rows of $W^\ast$) correspond to the principal components of the data signals. We note that this observation comes with a caveat; when the noise variance is high, it will dominate the contribution to the measurements $Y$ in Eq.~\eqref{eq:non_linear_system} as one would expect. Hence, the measurements and the signal will have low mutual information even under the optimal measurement matrix $W^\ast$. 

Our assumptions are notably different from prior results in autoencoding drawing connections with PCA. Prior results show that linear encoding \textit{and} decoding in a standard autoencoder recovers the principal components of the data (Eq.~(3) in \cite{bourlard1988auto}, Eq.~(1) in \cite{baldi1989neural}). In contrast, Theorem~\ref{thm:pca} is derived from variational principles and \emph{does not assume linear decoding}. 

In general, the behaviors of UAE and PCA can be vastly different. As noted in prior work~\citep{barber2003algorithm,weiss2007learning}, the principal components may not be the the most informative low-dimensional projections for recovering the original high-dimensional data back from its projections. A UAE, on the other hand, is explicitly designed to preserve as much information as possible (see Eq.~\eqref{eq:cond_entropy}). We illustrate the differences in a synthetic experiment in Figure~\ref{fig:gmm}. The true data distribution is an equiweighted mixture of two Gaussians stretched along orthogonal directions. We sample $100$ points (black) from this mixture and consider two dimensionality reductions. In the first case, we project the data on the first principal component (blue points on magenta line). This axis captures a large fraction of the variance in the data but collapses data sampled from the bottom right Gaussian in a narrow region. The projections of the data on the UAE axis (red points on green line) are more spread out.
This suggest that recovery is easier, even if doing so increases the total variance in the projected space compared to PCA.

\section{EXPERIMENTS}\label{sec:exp}

\subsection{Statistical compressed sensing}
We perform compressed sensing on three datasets:  MNIST~\citep{lecun2010mnist}, Omniglot~\citep{lake2015human}, and CelebA dataset~\citep{liu2015faceattributes}, with extremely low number of measurements $m\in \{2,5,10,25,50, 100\}$. 
We discuss the MNIST and Omniglot datasets here since they have a similar setup. To save space, results on the CelebA dataset are deferred to Appendix~\ref{app:celebA}. Every image in MNIST and Omniglot has a dimensionality of $28 \times 28$.
In all our experiments, we assume a Gaussian noise model with $\sigma=0.1$. We evaluated UAE against:

\begin{itemize}[leftmargin=*]
\item \textbf{LASSO} decoding with random Gaussian matrices. 
The MNIST and Omniglot datasets are reasonably sparse in the canonical pixel basis, and hence, we did not observe any gains after applying
Discrete Cosine Transform
and Daubechies-1 Wavelet Transform.

\item \textbf{CS-VAE}. This approach to compressed sensing was proposed by \citet{bora2017compressed} and learns a latent variable generative model over the observed variables $X$ and the latent variables $Z$. 
Such a model defines a mapping $G: \mathbb{R}^k \rightarrow \mathbb{R}^n$ from $Z$ to $X$, which is given by either the mean function of the observation model for a VAE or the forward deterministic mapping to generate samples for a GAN. We use VAEs in our experiments.
Thereafter, using a classic acquisition matrix satisfying a generalized Restricted Eigenvalue Condition (say $W$) (\eg, random Gaussian matrices), the reconstruction $\hat{x}$ for any datapoint is given as:
$
\hat{x}=G(\mathrm{arg} \min_{z} \Vert y - W G(z) \Vert_2)
$. 
Intuitively, this procedure seeks the latent vector $z$ such that the corresponding point on the range of $G$ can best approximate the measurements $y$ under the mapping $W$. 
We used the default parameter settings and architectures proposed in \citet{bora2017compressed}.  

\item \textbf{RP-UAE.} To independently evaluate the effect of variational decoding, this ablation baseline encodes the data using Gaussian random projections (RP) and trains the decoder based on the UAE objective. Since LASSO and CS-VAE both use an RP encoding, the differences in performance would arise only due to the decoding procedures.
\end{itemize}

The UAE decoder and the CS-VAE encoder/decoder are multi-layer perceptrons consisting of two hidden layers with $500$ units each.
For a fair comparison with random Gaussian matrices, the UAE encoder is linear. 
Further, we perform $\ell_2$ regularization on the norm of $W$. This helps in generalization to test signals outside the train set and is equivalent to solving the Lagrangian of a constrained UAE objective:
\begin{align*}
\max_{\theta, \phi}\mathbb{E}_{Q_\phi(X,Y)}\left[\log p_\theta(x\vert y)\right] 
\text{ subject to }
\Vert W \Vert_F \leq k.
\end{align*}
The Lagrangian parameter is chosen by line search on the above objective. 
The constraint ensures that UAE does not learn encodings $W$ that trivially scale the measurement matrix to overcome noise. For each $m$, we choose $k$ to be the expected norm of a random Gaussian matrix of dimensions $n \times m$ for fair comparisons with other baselines. In practice, the norm of the learned $W$ for a UAE is much smaller than those of random Gaussian matrices suggesting that the observed performance improvements are non-trivial. 

\textbf{Results.} The $\ell_2$ reconstruction errors on the standard test sets
are shown in Figure~\ref{fig:l2_reconstr}. For both datasets, we observe that UAE drastically outperforms both LASSO and CS-VAE for all values of $m$ considered. LASSO (blue curves) is unable to reconstruct with such few measurements. The CS-VAE (red) error decays much more slowly compared to UAE as $m$ grows. Even the RP-UAE baseline (yellow), which trains the decoder keeping the encoding fixed to a random projection, outperforms CS-VAE. 
Jointly training the encoder and the decoder using the UAE objective (green) exhibits the best performance.  
These results are also reflected qualitatively for the reconstructed test signals shown in Figure~\ref{fig:reconstr} for $m=25$ measurements. 

\begin{table*}[t]
\centering
  \caption{PCA vs. UAE. Average test classification accuracy for the MNIST dataset. 
  }
  \label{tab:pca_uae}
  \centering
\scriptsize
  \begin{tabular}{|c|c|c|c|c|c|c|c|c|c|c|}
    \toprule
	Dimensions & Method &  kNN & DT & RF & MLP & AdaB & NB & QDA & SVM \\
    \midrule
    2 & PCA & 0.4078	&	0.4283	&	0.4484	&	0.4695	&	0.4002	&	0.4455	&	0.4576	&	0.4503	
\\
      & UAE &\textbf{ 0.4644}	&	\textbf{0.5085}	&	\textbf{0.5341}	&	\textbf{0.5437}	&	\textbf{0.4248}	&	\textbf{0.5226}	&	\textbf{0.5316}	&	\textbf{0.5256}	
\\
    \midrule
     5 & PCA & 0.7291	&	0.5640	&	0.6257	&	0.7475	&	0.5570	&	0.6587	&	0.7321	&	0.7102	
\\
      & UAE & \textbf{0.8115}	&	\textbf{0.6331}	&	\textbf{0.7094}	&	\textbf{0.8262}	&	\textbf{0.6164}	&	\textbf{0.7286	}&	\textbf{0.7961}	&	\textbf{0.7873}
 \\
      \midrule
     10 & PCA & 0.9257	&	\textbf{0.6354}	&	0.6956	&	0.9006	&	0.7025	&	0.7789	&	0.8918	&	0.8440	\\
      & UAE &\textbf{ 0.9323}	&	0.5583	&	\textbf{0.7362}	&	\textbf{0.9258}	&	\textbf{0.7165}	&	\textbf{0.7895}	&	\textbf{0.9098}	&	\textbf{0.8753}	
\\
   \midrule
     25 & PCA & \textbf{0.9734}	&	\textbf{0.6382}	&	0.6889	&	0.9521	&	0.7234	&	\textbf{0.8635}	&	0.9572	&	0.9194	\\
      & UAE & 0.9730	&	0.5407	&	\textbf{0.7022}	&	\textbf{0.9614}	&	\textbf{0.7398}	&	0.8306	&	\textbf{0.9580}	&	\textbf{0.9218}	\\
     \midrule
     50 & PCA & 0.9751	&	\textbf{0.6381}	&	0.6059	&	0.9580	&	\textbf{0.7390}	&	\textbf{0.8786}	&	0.9632	&	0.9376	
\\
      & UAE & \textbf{0.9754}	&	0.5424	&	\textbf{0.6765}	&	\textbf{0.9597}	&	0.7330	&	0.8579	&	\textbf{0.9638}	&	\textbf{0.9384}	
\\
   \midrule
     100 & PCA & \textbf{0.9734}	&	0.6380	&	0.4040	&	0.9584	&	0.7136	&	0.8763	&	0.9570	&	0.9428	
\\
      & UAE & 0.9731	&	\textbf{0.6446}	&	\textbf{0.6241}	&	\textbf{0.9597}	&	\textbf{0.7170}	&	\textbf{0.8809}	&	\textbf{0.9595}	&	\textbf{0.9431}	
\\
    \bottomrule
  \end{tabular}
\end{table*}
\normalsize
\subsection{Transfer compressed sensing}

To test the generalization of the learned models to similar, unseen datasets, we consider the task of \textit{transfer} compressed sensing task introduced in \citet{dhar2018modeling}.

\textbf{Experimental setup.} We train the models on a source domain that is related to a target domain. Since the dimensions of MNIST and Omniglot images match, transferring from one domain to another requires no additional processing.
For UAE, we consider two variants. 
In UAE-SE, we used the encodings from the source domain and retrain the decoder on the target domain. 
For UAE-SD, we use source decoder and retrain the encoder on the target domain.

\textbf{Results.} The $\ell_2$ reconstruction errors
are shown in Figure~\ref{fig:l2_reconstr_transfer}. 
LASSO (blue curves) does not involve any learning, and hence its performance is same as Figure~\ref{fig:l2_reconstr}. 
The CS-VAE (red) performance degrades significantly in comparison, even performing worse than LASSO in some cases. 
The UAE based methods outperform these approaches and UAE-SE (green) fares better than UAE-SD (yellow). 
Qualitative differences are highlighted in Figure~\ref{fig:reconstr_transfer} for 
$m=25$ measurements. 

\subsection{Dimensionality reduction}
Dimensionality reduction is a common preprocessing technique for specifying features for classification. We compare PCA and UAE on this task. While Theorem~\ref{thm:pca} posits that the two techniques are equivalent in the regime of high noise given optimal UAE decodings, we set the noise as a hyperparameter based on a validation set to enable out-of-sample generalization.

\textbf{Setup.} 
We learn the principal components and UAE projections on the MNIST training set for varying number of dimensions. We then learn classifiers based on the these projections. Again, we use a linear encoder for the UAE for a fair evaluation.
Since the inductive biases vary across different classifiers, we considered $8$ commonly used classifiers: k-Nearest Neighbors (kNN), Decision Trees (DT), Random Forests (RF), Multilayer Perceptron (MLP), AdaBoost (AdaB), Gaussian Naive Bayes (NB), Quadratic Discriminant Analysis (QDA), and Support Vector Machines (SVM) with a linear kernel. 

\textbf{Results.} The performance of the PCA and UAE feature representations for different number of dimensions is shown in Table~\ref{tab:pca_uae}. We find that UAE outperforms PCA in a majority of the cases. Further, this trend is largely consistent across classifiers. The improvements are especially high when the number of dimensions is low, suggesting the benefits of UAE as a dimensionality reduction technique for classification. 

\section{RELATED WORK}\label{sec:discussion}

In this section, we contrast uncertainty autoencoders with related works in autoencoding, compressed sensing, and mutual information maximization.

\textbf{Autoencoders.} To contrast uncertainty autoencoders with other commonly used autoencoding schemes, consider a UAE with a Gaussian observation model with fixed isotropic covariance for the decoder of all the autoencoding objectives we discuss subsequently. 
The UAE objective can be simplified as:
\begin{align*}
\min_{\theta, \phi}\mathbb{E}_{x,y \sim Q_\phi(X,Y)}\left[\Vert x - g_\theta(y)\Vert_2^2\right]
\end{align*}

\textbf{Standard Autoencoder.} If we assume no measurement noise (\textit{i.e.}, $\epsilon = 0$) and assume the observation model $P_{\theta}(X\vert Y)$ to be a Gaussian with mean $g_\theta(Y)$ and a fixed isotropic $\Sigma$, then the UAE objective reduces to minimizing the mean squared error between the true and recovered datapoint:
\begin{align*}
\min_{\theta, W, \psi}\mathbb{E}_{x \sim Q_{\mathrm{data}}(X)}\left[\Vert x - g_\theta(Wf_\psi(x))\Vert_2^2\right]
\end{align*}
This special case of a UAE corresponds to a standard autoencoder~\citep{bengio2009learning} where the measurements $Y$ signify a hidden representation for $X$. However, this case lacks the interpretation of an implicit generative model since the assumptions of Theorem~\ref{thm:markov_chain} do not hold.

\textbf{Denoising Autoencoders.} A DAE~\citep{vincent2008extracting} 
adds noise at the level of the input datapoint $X$ to learn robust representations. For a UAE, the noise model is defined at the level of the compressed measurements. 
Again, with the assumptions of a Gaussian decoder, the DAE objective can be expressed as:
\begin{align*}
\min_{\theta, W, \psi}\mathbb{E}_{x \sim Q_{\mathrm{data}}(X), \tilde{x} \sim C(\tilde{X} \vert x)}\left[\Vert x - g(Wf_\psi(\tilde{x}))\Vert_2^2\right]
\end{align*}
where $C(\cdot \vert X)$ is some predefined noise corruption model. 
Similar to Theorem~\ref{thm:markov_chain}, a DAE also learns an implicit model of the data distribution~\citep{bengio2013generalized,alain2014regularized}.

\textbf{Variational Autoencoders.} A VAE~\citep{kingma-iclr2014,rezende2014stochastic} explicitly learns a latent variable model $P_\theta(X,Y)$ for the dataset. The learning objective is a variational lower bound to the marginal log-likelihood assigned by the model to the data $\mathcal{X}$, which notationally  corresponds to  $\mathbb{E}_{ Q_{\mathrm{data}}(X)}[\log P_\theta(x)]$. The variational objective that maximizes this quantity can be simplified as:
\begin{align*}
\min_{\theta, \phi}& \quad \mathbb{E}_{x,y \sim Q_\phi(X,Y)}\left[\Vert x - g_\theta(y)\Vert_2^2\right] \\
&+ \mathbb{E}_{x \sim Q_\mathrm{data}}\left[KL(Q_\phi(Y \vert x), P(Y))\right]
\end{align*}

The learning objective includes a reconstruction error term, akin to the UAE objective. Crucially, it also includes a regularization term to minimize the $\mathrm{KL}$ divergence of the variational posterior
over $Y$
with a prior distribution over $Y$.
A key difference is that a UAE does not explicitly need to model the prior distribution over $Y$. 
On the downside, a VAE can perform efficient ancestral sampling while a UAE requires running relatively expensive Markov Chains to obtain samples.

Recent works have attempted to unify the variants of variational autoencoders through the lens of mutual information~\citep{alemi2017information,zhao2018information,chen2016variational}. These works also highlight scenarios where the VAE can learn to ignore the latent code in the presence of a strong decoder thereby affecting the reconstructions to attain a lower KL loss. 
One particular variant, the $\beta$-VAE, weighs the additional KL
regularization term with a positive factor $\beta$ and can effectively learn disentangled representations~\citep{higgins2016beta,zhao2019infovae}. Although \citep{higgins2016beta} does not consider this case, the UAE can be seen as a $\beta$-VAE with $\beta=0$.

To summarize, our uncertainty autoencoding formulation provides a combination of unique desirable properties for representation learning that are absent in prior autoencoders. 
As discussed, a UAE defines an implicit generative model without specifying a prior (Theorem~\ref{thm:markov_chain}) even under realistic conditions (Corollary~\ref{thm:markov_chain_gaussian}; unlike DAEs) and has rich connections with PCA even for non-linear decoders (Theorem~\ref{thm:pca}; unlike any kind of existing autoencoder).

\textbf{Generative modeling and compressed sensing.} The closely related works of \cite{bora2017compressed,dhar2018modeling} also use generative models for compressed sensing. As highlighted in Section~\ref{sec:exp}, their approach is radically different from UAE. Similar to \cite{bora2017compressed}, a UAE learns a data distribution. However, in doing so, it additionally learns an acquisition/encoding function and a recovery/decoding function, unlike \cite{bora2017compressed,dhar2018modeling} which rely on generic random matrices and $\ell_2$ decoding. The cost of implicit learning in a UAE is that some of its inference capabilities, such as likelihood evaluation and sampling, are intractable or require running Markov chains. However, these inference queries are orthogonal to compressed sensing. Finally, our decoding is amortized and scales to large datasets, unlike \citet{bora2017compressed,dhar2018modeling} which solve an independent optimization problem for each test datapoint.

\paragraph{Mutual information maximization.} The principle of mutual information maximization, often referred to as InfoMax in prior work, was first proposed for learning encodings for communication over a noisy channel~\citep{linsker1989generate}. 
The InfoMax objective has also been applied for statistical compressed sensing for learning both linear and non-linear encodings~\citep{weiss2007learning,carson2012communications,wang2014nonlinear}. Our work differs from these existing frameworks in two fundamental ways. First, we optimize for a tractable variational lower bound to the MI that which allows our method to scale to high-dimensional data. Second, we learn an amortized~\citep{gershman2014amortized,shu2018amortized} decoder in addition to the encoder that sidesteps expensive, per-example optimization for the test datapoints.

Further, we improve upon the \textit{IM algorithm} proposed originally for variational information maximization~\citep{barber2003algorithm}. While the IM algorithm proposes to optimize the lower bound on the mutual information in alternating ``wake-sleep'' phases for optimizing the encoder (``wake") and decoder (``sleep") analogous to the expectation-maximization procedure used in \citet{weiss2007learning}, we optimize the encoder and decoder jointly using a single consistent objective leveraging recent advancements in gradient based variational stochastic optimization.

\section{CONCLUSION}
In this work, we presented uncertainty autoencoders (UAE), a framework for unsupervised representation learning via variational maximization of mutual information between an input signal and its latent representation. 
We presented connections of our framework with many related threads of research, in particular with respect to implicit generative modeling and principal component analysis. 
 Empirically, we showed that UAEs are a natural candidate for statistical compressed sensing, wherein we can learn the acquisition and recovery functions jointly.

In the future, it would be interesting to incorporate advancements in compressed sensing based on complex neural network architectures~\citep{mousavi2015deep,kulkarni2016reconnet,chang2017one,lu2018convcsnet,van2018compressed} within the UAE framework for real world applications, \textit{e.g.}, medical imaging.
Unlike the rich theory surrounding the compressed sensing of sparse signals, a similar theory surrounding generative model-based priors on the signal distribution is lacking. 
Recent works have made promising progress in developing a theory of SGD based recovery methods for nonconvex inverse problems, which continues to be an exciting direction for future work~\citep{bora2017compressed,hand2017global,dhar2018modeling,liu2018convergence}.

\section*{Acknowledgements}
This research was supported by NSF (\#1651565, \#1522054, \#1733686), ONR (N00014-19-1-2145), AFOSR (FA9550-19-1-0024), and FLI.
AG is supported by a Microsoft Research Ph.D. fellowship and a Stanford Data Science scholarship.
We are thankful to Daniel Levy and Yang Song for helpful discussions on a proof and Kristy Choi, Manik Dhar, Neal Jean, and Ben Poole for helpful comments. 

\bibliographystyle{unsrt}
\bibliography{refs}
\newpage
\appendix
\onecolumn
\section{Proofs of Theoretical Results}\label{app:proof}
\subsection{Proof of Theorem~\ref{thm:markov_chain}}\label{app:markov_chain}

\begin{proof}
We can rewrite the UAE objective in Eq.~\eqref{eq:uae_obj} as:
\begin{align}
\mathbb{E}_{Q_\phi(X,Y)}\left[\log p_\theta(x\vert y) 
\right]
&= \mathbb{E}_{Q_\phi(Y)}\left[\int q_\phi(x\vert y)\log p_\theta(x\vert y) \mathrm{d}x\right] \\
&= -H_\phi(X \vert Y) - \mathbb{E}_{Q_\phi(Y)}
\left[\mathrm{KL}(Q_\phi(X\vert y)\Vert  P_\theta(X\vert y))\right].
\end{align}
The $\mathrm{KL}$-divergence is non-negative and minimized when its argument distributions are identical. Hence, for a fixed optimal value of $\theta=\theta^\ast$, if there exists a $\phi$ in the space of encoders being optimized that satisfies:
\begin{align}\label{eq:posterior_match}
p_{\theta^\ast}(X\vert Y) = q_\phi(X\vert Y)
\end{align}
for all $X,Y$ with $p_\theta^\ast(Y) \neq 0$, then it corresponds to the optimal encoder, \ie,
\begin{align}\label{eq:posterior_match_2}
\phi=\phi^\ast.
\end{align}

For any value of $\phi$, we know the following Gibbs chain converges to $Q_\phi(X,Y)$ if the chain is ergodic:
\begin{align}
y^{(t)} &\sim Q_{\phi}(Y \vert x^{(t)})\label{eq:true_markov_chain_1}\\
x^{(t+1)} &\sim Q_{\phi}(X \vert y^{(t)})\label{eq:true_markov_chain_2}.
\end{align}
Substituting the results from Eqs.~(\ref{eq:posterior_match}-\ref{eq:true_markov_chain_2}) in the Markov chain transitions in Eqs.~(\ref{eq:markov_chain_1}, \ref{eq:markov_chain_2}) finishes the proof.
\end{proof}

\subsection{Proof of Corollary~\ref{thm:markov_chain_gaussian}}\label{app:markov_chain_gaussian}
\begin{proof}
By using earlier results (Proposition 2 in \cite{roberts2006harris}), we need to show that the Markov chain defined in Eqs.~\eqref{eq:markov_chain_1}-\eqref{eq:markov_chain_2} is $\Phi$-irreducible with a Gaussian noise model.\footnote{Note that the symbol $\Phi$ here is different from the parameters denoted by little $\phi$ used in the rest of the paper.} That is, there exists a measure such that there is a non-zero probability of transitioning from every set of non-zero measure to every other such set defined on the same measure using this Markov chain. 

Consider the Lebesgue measure. Formally, given any $(x, y)$ and $(x',y')$ such that the density $q(x,y) > 0$ and $q(x',y')>0$ for the Lebesgue measure, we need to show that the probability density of transitioning $q(x',y'\vert x,y)>0$.

(1) Since $q(y\vert x)>0$ for all $x \in \mathbb{R}^n$, $y \in \mathbb{R}^m$ (by Gaussian noise model assumption), we can use Eq.~\eqref{eq:markov_chain_1} to transition from $(x,y)$ to $(x, y')$ with non-zero probability.  
 
(2) Next, we claim that the transition probability $q(x' \vert y)$ is non-negative for all $x', y$. By Bayes rule, we have:
\[
                  q(x'\vert y) = \frac{q(y\vert x') q(x')}{q(y)}.
\]

Since $q(x,y)>0$ and $q(x',y')>0$, the marginals $q(y)$ and $q(x')$ are positive. Again, $q(y\vert x')>0$ for all $x' \in \mathbb{R}^n$, $y \in \mathbb{R}^m$ by the Gaussian noise model assumption.  Hence, $q(x'\vert y)$  is positive. Finally, using the optimality assumption for the posteriors $p(x'\vert y)$ matching $q(x'\vert y)$ for all $x', y'$, we can use Eq.~\eqref{eq:markov_chain_2} to transition from $(x,y')$ to $(x',y')$ with non-zero probability.

From (1) and (2), we see that there is a non-zero probability of transitioning from $(x,y)$ to $(x',y')$. Hence, under the assumptions of the corollary the Markov chain in Eqs.~(\ref{eq:markov_chain_1}, \ref{eq:markov_chain_2}) is ergodic.

\end{proof}

\subsection{Proof of Theorem~\ref{thm:pca}}\label{app:pca}

\begin{proof}
Under an optimal decoder, the model posterior $P_\theta(X \vert Y)$ matches the true posterior $Q_\phi(X \vert Y)$ and hence, the UAE objective can be simplified as:
\begin{align}
\mathbb{E}_{Q_\phi(X,Y)}[\log q_\phi(x \vert y)]&=\mathbb{E}_{Q_\phi(X,Y)}[\log q_\phi(x, y) - \log q_\phi(y)]\nonumber \\
&=-H(X) - \mathbb{E}_{Q_{\mathrm{data}}(X)}[H(Y \vert x)] - \mathbb{E}_{Q_\phi(X,Y)}[\log q_\phi(y)].
\end{align}

The first term corresponds to the negative of the data entropy, is independent of $\phi$ and $\sigma$, and hence it can be removed. 
For the second term, note that $Y\vert x$ is a normal distributed random variable and hence its entropy is given by a constant $\frac{1}{2} \log 2\pi e\sigma^2$.
Only the third term depends on $\phi$. 

Removing the data entropy term since it is a constant independent of both $\phi$ and $\sigma$, we can define a modified objective $M(W, \mathcal{D}, \sigma)$ as:

\begin{align}
M(W, \mathcal{D}, \sigma) := \mathbb{E}_{Q_\phi(X,Y)}[\log q_\phi(y)] + \frac{1}{2} \log 2\pi e \sigma^2. 
\end{align}
As $\sigma \to \infty$, the optimal encodings maximizing the mutual information can be specified as:
\begin{align}\label{eq:log_marginal}
W^\ast =  \lim_{\sigma \to \infty} \arg\max_W - M(W, \mathcal{D}, \sigma).
\end{align}

We can lower-bound $M(W,\mathcal{D}, \sigma)$ using Jensen's inequality:

\begin{align}\label{eq:marginal_ub}
M(W, \mathcal{D}, \sigma)
&= \mathbb{E}_{Q_\phi(X,Y)}[\log \mathbb{E}_{x_j \sim Q_{\mathrm{data}}(X)}\left[q_\phi(y \vert x_j )\right]] + \frac{1}{2} \log 2\pi e \sigma^2\nonumber\\
&= \frac{1}{\vert\mathcal{D}\vert}\sum_{x_i\in \mathcal{D}}\mathbb{E}_{Q_\phi(Y \vert X)}\left[\log\frac{1}{\vert\mathcal{D}\vert}\sum_{x_j\in \mathcal{D}} q_\phi(y \vert x_j )\right]+\frac{1}{2} \log 2\pi e \sigma^2\nonumber\\
&\geq \frac{1}{\vert\mathcal{D}\vert}\sum_{x_i\in \mathcal{D}}\mathbb{E}_{Q_\phi(Y \vert X)}\left[\sum_{x_j\in \mathcal{D}} \frac{1}{\vert\mathcal{D}\vert}\log  q_\phi(y \vert x_j ) \right] +\frac{1}{2} \log 2\pi e \sigma^2\nonumber\\
&:= C(W, \mathcal{D}, \sigma)
\end{align}
where we have used the fact that the data distribution is uniform over the entire dataset (by assumption).

Finally, we denote the non-negative slack term for the above inequality as $S(W, \mathcal{D}, \sigma)$ such that:
\begin{align}\label{eq:decomp}
 M(W, \mathcal{D}, \sigma) &= C(W, \mathcal{D}, \sigma) + S(W, \mathcal{D}, \sigma).  
\end{align}

\textit{Overview of proof strategy:} We will first simplify expressions for the lower bound $C(W, \mathcal{D}, \sigma)$ and slack term $S(W, \mathcal{D}, \sigma)$.
Then, we will show that as $\sigma \to \infty$, the ratio of the slack term and the lower bound converges pointwise to $0$ and hence, the lower bound is arbitrarily close to $M(W, \mathcal{D}, \sigma)$ in this regime for a fixed $W$.
Further, we will show that the convergence is uniform in $W$.
Finally, we will note that the optimal encodings $W^\ast$ for the lower bound correspond to the stated expressions for $W$ in the proof statement.

As a first step, we consider simplifications of the lower bound and the slack term.

\textbf{Lower bound:}  $C(W, \mathcal{D}, \sigma)$
\begin{align}\label{eq:c_simp}
C(W, \mathcal{D}, \sigma)&= \frac{1}{\vert\mathcal{D}\vert}\sum_{x_i\in \mathcal{D}}\mathbb{E}_{\epsilon \sim \mathcal{N}(0, \sigma^2I)}\left[\frac{1}{\vert\mathcal{D}\vert}\sum_{x_j\in \mathcal{D}}[\log q_\phi(Wx_i + \epsilon \vert x_j )]\right]+\frac{1}{2} \log 2\pi e \sigma^2\nonumber\\
&= \frac{1}{\vert\mathcal{D}\vert^2}\sum_{x_i, x_j\in \mathcal{D}}\mathbb{E}_{\epsilon \sim \mathcal{N}(0, \sigma^2I)}[\log q_\phi(Wx_i + \epsilon \vert x_j )]+\frac{1}{2} \log 2\pi e \sigma^2\nonumber\\
&= -\frac{1}{\vert\mathcal{D}\vert^2}\sum_{x_i, x_j\in \mathcal{D}}\mathbb{E}_{\epsilon \sim \mathcal{N}(0, \sigma^2I)}\left[ \frac{(Wx_i + \epsilon-Wx_j)^T(Wx_i + \epsilon-Wx_j)}{2\sigma^2} +\frac{1}{2}\log 2 \pi \sigma^2 \right]+\frac{1}{2} \log 2\pi e \sigma^2\nonumber\\
&= -\frac{1}{\vert\mathcal{D}\vert^2}\sum_{x_i, x_j\in \mathcal{D}}\mathbb{E}_{\epsilon \sim \mathcal{N}(0, \sigma^2I)}\left[ \frac{(Wx_i-Wx_j)^T(Wx_i -Wx_j) + 2 \epsilon^T(Wx_i -Wx_j) + \epsilon^T\epsilon}{2\sigma^2}  \right] +  \frac{1}{2}\nonumber\\
&= -\frac{1}{\vert\mathcal{D}\vert^2}\sum_{x_i, x_j\in \mathcal{D}} \left(\frac{(Wx_i-Wx_j)^T(Wx_i -Wx_j)}{2\sigma^2} +  \frac{1}{2}\right) +  \frac{1}{2}\nonumber \\
&= -\frac{1}{\vert\mathcal{D}\vert^2}\sum_{x_i, x_j\in \mathcal{D}} \left(\frac{(Wx_i-Wx_j)^T(Wx_i -Wx_j)}{2\sigma^2} \right).
\end{align}

\textbf{Slack:}  $S(W, \mathcal{D}, \sigma)$

\begin{align}
S(W, \mathcal{D}, \sigma) &= -C(W, \mathcal{D}, \sigma) + M(W, \mathcal{D}, \sigma)\nonumber\\
&=-\frac{1}{\vert\mathcal{D}\vert}\sum_{x_i\in \mathcal{D}}\mathbb{E}_{Q_\phi(Y \vert X)}\left[\frac{1}{\vert\mathcal{D}\vert}\sum_{x_j\in \mathcal{D}}\log q_\phi(y \vert x_j )\right] +\frac{1}{\vert\mathcal{D}\vert}\sum_{x_i\in \mathcal{D}}\mathbb{E}_{Q_\phi(Y \vert X)}\left[\log\frac{1}{\vert\mathcal{D}\vert}\sum_{x_j\in \mathcal{D}} q_\phi(y \vert x_j )\right]\nonumber\\
&= -\frac{1}{\vert\mathcal{D}\vert}\sum_{x_i\in \mathcal{D}}\mathbb{E}_{Q_\phi(Y \vert X)}\left[\mathbb{E}_{Q_\mathrm{data}(X)}\left[\log \frac{q_\phi(y , x_j )}{q_\mathrm{data}(x_j)}\right]\right] +\frac{1}{\vert\mathcal{D}\vert}\sum_{x_i\in \mathcal{D}}\mathbb{E}_{Q_\phi(Y \vert X)}\left[\log q_\phi(y)\right]\nonumber\\
&= \frac{1}{\vert\mathcal{D}\vert}\sum_{x_i\in \mathcal{D}}\mathbb{E}_{Q_\phi(Y \vert X)}\left[\mathbb{E}_{Q_\mathrm{data}(X)}\left[\log \frac{q_\mathrm{data}(x_j)}{q_\phi(x_j \vert y)q_\phi(y)}\right]\right] +\frac{1}{\vert\mathcal{D}\vert}\sum_{x_i\in \mathcal{D}}\mathbb{E}_{Q_\phi(Y \vert X)}\left[\log q_\phi(y)\right]\nonumber\\
&= \frac{1}{\vert\mathcal{D}\vert}\sum_{x_i\in \mathcal{D}}\mathbb{E}_{Q_\phi(Y \vert X)}\left[KL\left(Q_\mathrm{data}(X),Q_\phi(X \vert y)\right)\right] \nonumber\\
&=- \frac{1}{\vert\mathcal{D}\vert^2}\sum_{x_i, x_j\in \mathcal{D}}\mathbb{E}_{Q_\phi(Y \vert X)}\left[  \log q_\phi(x_j \vert y) + \log \vert \mathcal{D} \vert \right]\nonumber\\
&= - \log \vert \mathcal{D}\vert -\frac{1}{\vert\mathcal{D}\vert^2}\sum_{x_i, x_j\in \mathcal{D}} \mathbb{E}_{\epsilon \sim \mathcal{N}(0, \sigma^2I)}\left[\log q_\phi(x_j \vert W x_i + \epsilon)\right]
\end{align}

We can simplify the posteriors $Q_\phi(x_j \vert W x_i + \epsilon)$ as:
\begin{align}\label{eq:true_posterior}
&Q_\phi(x_j \vert W x_i + \epsilon) 
= \frac{Q_\phi(x_j , W x_i + \epsilon)}{Q_\phi(W x_i + \epsilon)}\nonumber\\
 &= \frac{Q_\phi(x_j) Q_\phi(W x_i + \epsilon \vert  x_j)}{\sum_{x_k \in \mathcal{D}}Q_\phi(W x_i + \epsilon \vert x_k)Q_\phi(x_k)}
= \frac{\exp\left(-\nicefrac{(W(x_i-x_j) + \epsilon)^T(W(x_i-x_j) + \epsilon)}{2 \sigma^2}\right)}{\sum_{x_k \in \mathcal{D}} \exp\left(-\nicefrac{(W(x_i-x_k) + \epsilon)^T(W(x_i-x_k) + \epsilon)}{2 \sigma^2}\right)}
\end{align}
where we have used the fact that the data distribution is uniform and the decoder is isotropic Gaussian. 

Substituting the above expression for the slack term:
\begin{align}\label{eq:s_simp}
S(W, \mathcal{D}, \sigma) &= - \log \vert \mathcal{D}\vert -\frac{1}{\vert\mathcal{D}\vert^2}\sum_{x_i, x_j\in \mathcal{D}}\mathbb{E}_{\epsilon \sim \mathcal{N}(0, \sigma^2I)}\bigg[ \left(-\frac{(W(x_i-x_j) + \epsilon)^T(W(x_i-x_j) + \epsilon)}{2 \sigma^2}\right) \nonumber\\
& - \log \sum_{x_k \in \mathcal{D}} \exp\left(-\frac{(W(x_i-x_k) + \epsilon)^T(W(x_i-x_k) + \epsilon)}{2 \sigma^2}\right)\bigg] \nonumber\\
&=  \frac{1}{\vert\mathcal{D}\vert^2}\sum_{x_i, x_j\in \mathcal{D}}\bigg[ \frac{(Wx_i-Wx_j)^T(Wx_i -Wx_j)}{2\sigma^2} +  \frac{1}{2} \nonumber\\
&+ \underbrace{\mathbb{E}_{\epsilon \sim \mathcal{N}(0, \sigma^2I)} \left(\log \sum_{x_k \in \mathcal{D}} \exp\left(-\frac{(W(x_i-x_k) + \epsilon)^T(W(x_i-x_k) + \epsilon)}{2 \sigma^2}\right) - \log \vert \mathcal{D}  \vert\right)}_{\int \gamma(\sigma, \epsilon, W,\mathcal{D}, x_i)\mathrm{d}\epsilon} \bigg].
\end{align}

For any fixed $x_i, W$, the integrand $\gamma(\sigma, \epsilon, W,\mathcal{D}, x_i)$ in Eq.~\eqref{eq:s_simp} can be seen as a sequence of functions indexed by $\sigma$. 
We next make the claim that dominated convergence in $\epsilon$ holds for this sequence for all $x_i, W$.
We show so by first observing that $\gamma(\sigma, \epsilon, W,\mathcal{D}, x_i)$ converges pointwise to a constant (=0) as $\sigma \to \infty$ and thereafter deriving integrable upper and lower bounds for $\gamma(\sigma, \epsilon, W,\mathcal{D}, x_i)$ below that are independent of $\sigma$.

For the upper bound, we note that:
\begin{align}
\log\sum_{x_j\in \mathcal{D}}\exp\left(- \frac{(Wx_i + \epsilon-Wx_j)^T(Wx_i + \epsilon-Wx_j)}{2\sigma^2}\right) & \leq \max_{x_j\in \mathcal{D}} - \frac{(Wx_i + \epsilon-Wx_j)^T(Wx_i + \epsilon-Wx_j)}{2\sigma^2} + \log \vert \mathcal{D} \vert \nonumber \\
&\leq \log \vert \mathcal{D} \vert.
\end{align}
This gives an upper bound on the integrand $\gamma(\sigma, \epsilon, W,\mathcal{D}, x_i)$:
\begin{align}
\gamma(\sigma, \epsilon, W,\mathcal{D}, x_i) &\leq - \frac{1}{\sqrt{2 \pi \sigma^2}}\exp\left(-\frac{\epsilon^T\epsilon}{2\sigma^2}\right) \left [\log \vert \mathcal{D}\vert-\log \vert \mathcal{D}\vert\right ] \nonumber \\
&= 0.
\end{align}

For the lower bound, we note that:
\begin{align}
\log\sum_{x_j\in \mathcal{D}}\exp\left(- \frac{(Wx_i + \epsilon-Wx_j)^T(Wx_i + \epsilon-Wx_j)}{2\sigma^2}\right)&\geq \max_{x_j\in \mathcal{D}} \left(- \frac{(Wx_i + \epsilon-Wx_j)^T(Wx_i + \epsilon-Wx_j)}{2\sigma^2}\right)\nonumber\\
&= - \min_{x_j\in \mathcal{D}} \left(\frac{(Wx_i + \epsilon-Wx_j)^T(Wx_i + \epsilon-Wx_j)}{2\sigma^2}\right).
\end{align}.

Hence, we have the following lower bound:
\begin{align}
\gamma(\sigma, \epsilon, W,\mathcal{D}, x_i) &\geq - \frac{1}{\sqrt{2 \pi \sigma^2}}\exp\left(-\frac{\epsilon^T\epsilon}{2\sigma^2}\right) \left(\min_{x_j\in \mathcal{D}} \frac{(Wx_i + \epsilon-Wx_j)^T(Wx_i + \epsilon-Wx_j)}{2\sigma^2}-\log \vert \mathcal{D}\vert\right)\nonumber\\
&\geq -\frac{1}{\sqrt{\pi\epsilon^T\epsilon}}
\left(\frac{1}{\epsilon^T\epsilon}\left[\min_{x_j\in \mathcal{D}} (Wx_i + \epsilon-Wx_j)^T(Wx_i + \epsilon-Wx_j)\right] -2\log \vert \mathcal{D}\vert\right)\nonumber\\
&= -\frac{1}{\sqrt{\pi\epsilon^T\epsilon}}
\left(\frac{1}{\epsilon^T\epsilon}\left[\min_{x_j\in \mathcal{D}} (Wx_i -Wx_j)^T(Wx_i -Wx_j) + 2(Wx_i -Wx_j)^T\epsilon + \epsilon^T \epsilon\right] -2\log \vert \mathcal{D}\vert\right)\nonumber\\
&\geq -\frac{1}{\sqrt{\pi\epsilon^T\epsilon}}\left(\frac{4k_1^2 k_2^2}{\epsilon^T\epsilon} + \frac{4k_1k_2}{\sqrt{\epsilon^T\epsilon}} + 1-2\log \vert \mathcal{D}\vert\right)
\end{align}
where we used the inequalities $\exp(-1/z) \leq z^{3/2}, \exp(-1/z) \leq z^{1/2}$ for any $z>0$ in the second step (with $z=\nicefrac{2\sigma^2}{\epsilon^T\epsilon}$) and Cauchy-Schwarz for the last step (since $\Vert x\Vert_2 \leq k_1$ for all $x \in \mathcal{D}$, $\Vert W\Vert_F \leq k_2$ for some positive constants $k_1, k_2 \in \mathbb{R}^{+}$ by assumption).

Since both the upper and lower bounds for the integrand are independent of $\sigma$, dominated convergence holds for the third term in Eq.~\eqref{eq:s_simp}.

Consequently, we can evaluate limits to obtain a limiting ratio between the slack term and the lower bound:
\begin{align}
\lim_{\sigma \to \infty} \frac{S(W, \mathcal{D}, \sigma)}{C(W, \mathcal{D}, \sigma)} = 0
\end{align}
using the expressions derived in Eq.~\eqref{eq:c_simp} and Eq.~\eqref{eq:s_simp}, dominated convergence for interchanging limits and expectations, along with L'H\^opital's rule.

We can now rewrite Eq.~\eqref{eq:decomp} as:
\begin{align}
 M(W, \mathcal{D}, \sigma) &= C(W, \mathcal{D}, \sigma)\left(1 + \frac{S(W, \mathcal{D}, \sigma)}{C(W, \mathcal{D}, \sigma)}\right) .
\end{align}

By the $(\epsilon, \delta)$ definition of limit, we know that for any fixed $W$ that satisfies $\Vert W \Vert_F\leq k_2$ and $\forall \epsilon>0$, there exists a $\delta>0$ such that $\forall \sigma > \delta$, we have:
\begin{align}
\vert M(W, \mathcal{D}, \sigma) - C(W, \mathcal{D}, \sigma) \vert < \epsilon.
\end{align}

Next, we note that the slack term $S(W, \mathcal{D}, \sigma)$ is monotonic in $\sigma$ and converges pointwise for any fixed $W$ that satisfies $\Vert W \Vert_F\leq k_2$. 
\begin{align}
    \lim_{\sigma \to \infty} S(W, \mathcal{D}, \sigma) &= \lim_{\sigma \to \infty} M(W, \mathcal{D}, \sigma) - C(W, \mathcal{D}, \sigma) =0.
\end{align}

Using Dini's Theorem, this implies the convergence of the slack term is uniform in $W$ as $\sigma \to \infty$. Hence, for all $W$ that satisfy $\Vert W \Vert_F\leq k_2$ and $\forall \epsilon>0$, there exists a $\delta>0$ such that $\forall \sigma > \delta$, we have:
\begin{align}
\vert M(W, \mathcal{D}, \sigma) - C(W, \mathcal{D}, \sigma) \vert < \epsilon.
\end{align}

Since the $\arg\max$ operator preserves continuity (via Berge's maximum theorem) and is assumed to be identifiable, we conclude that $\forall W$ satisfying $\Vert W \Vert_F\leq k_2$ and $\forall \epsilon>0$, there exists a $\delta>0$ such that $\forall \sigma > \delta$, we have:
\begin{align}
\vert W^\ast - \arg\max_{W}\sum_{x_i, x_j\in \mathcal{D}}\left(Wx_i-Wx_j)^T(Wx_i -Wx_j)\right) \vert < \epsilon
\end{align}

which finishes the proof.
\end{proof}

\newpage
\section{Experimental details}\label{app:exp}
For MNIST, we use the train/valid/test split of $50,000/10,000/10,000$ images. For Omniglot, we use train/valid/test split of $23,845/500/8,070$ images. For CelebA, we used the splits as provided by ~\citet{liu2015faceattributes} on the dataset website. All images were scaled such that pixel values are between $0$ and $1$. We used the Adam optimizer with a learning rate of $0.001$ for all the learned models. For MNIST and Omniglot, we used a batch size of $100$. For CelebA, we used a batch size of $64$. Further, we implemented early stopping based on the best validation bounds after $200$ epochs for MNIST, $500$ epochs for Omniglot, and $200$ epochs for CelebA.

\subsection{Hyperparameters for compressed sensing on MNIST and Omniglot}

For both datasets, the UAE decoder used $2$ hidden layers of $500$ units each with ReLU activations. The encoder was a single linear layer with only weight parameters and no bias parameters. The encoder and decoder architectures for the \textit{VAE baseline} are symmetrical with $2$ hidden layers of $500$ units each and $20$ latent units. We used the \textit{LASSO baseline} implementation from sklearn and tuned the Lagrange parameter on the validation sets. For the baselines, we do $10$ random restarts with $1,000$ steps per restart and pick the reconstruction with best measurement error as prescribed in \citep{bora2017compressed}. Refer to \citep{bora2017compressed} for further details of the baseline implementations.

\begin{table}[t]
\centering
  \caption{Frobenius norms of the UAE encodings and random Gaussian projections for MNIST and Omniglot datasets.}
  \label{tab:mnist_norms}
  \centering
  \scriptsize
  \begin{tabular}{|c|c|c|c|}
    \toprule
	m & Random Gaussian Matrices & MNIST-UAE & Omniglot-UAE\\
    \midrule
     
     2 & 39.57 & 6.42 & 2.17\\
     5 & 63.15& 5.98 & 2.66\\
     10 & 88.98 & 7.24 & 3.50\\
     25 & 139.56 & 8.53 & 4.71\\
     50 & 198.28& 9.44 & 5.45\\
     100 & 280.25 &10.62& 6.02\\
    \bottomrule
  \end{tabular}
\end{table}

Table~\ref{tab:mnist_norms} shows the average norms for the random Gaussian matrices used in the baselines and the learned UAE encodings. The lower norms for the UAE encodings suggest that the UAE baseline is not trivially overcoming noise by increasing the norm of $W$.

\subsection{Hyperparameters for dimensionality reduction}
For PCA and each of the classifiers, we used the standard implementations in sklearn with default parameters and the following exceptions: 
\begin{itemize}
\item KNN: n\_neighbors = 3
\item DT: max\_depth = 5
\item RF: max\_depth = 5,  n\_estimators = 10, max\_features = 1
\item MLP: alpha=1
\item SVC: kernel=linear, C=0.025
\end{itemize}

\subsection{Statistical compressed sensing on CelebA dataset}\label{app:celebA}

\begin{figure*}[t]
\centering
\includegraphics[width=0.5\textwidth]{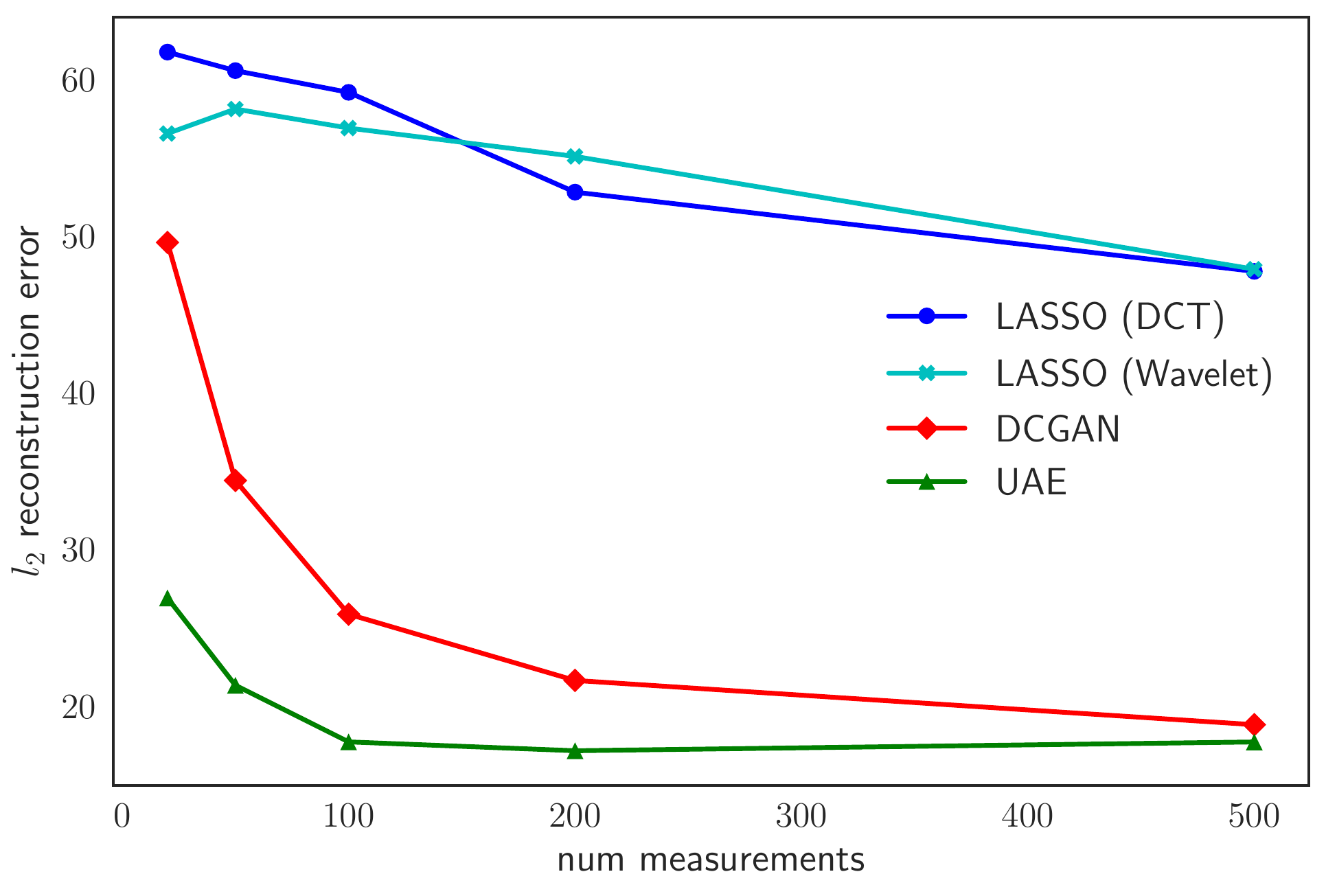}
\caption{Test $\ell_2$ reconstruction error (per image) for compressed sensing on CelebA.}\label{fig:celeba}
\end{figure*}

\begin{figure*}[t]

\centering
\includegraphics[width=\textwidth]{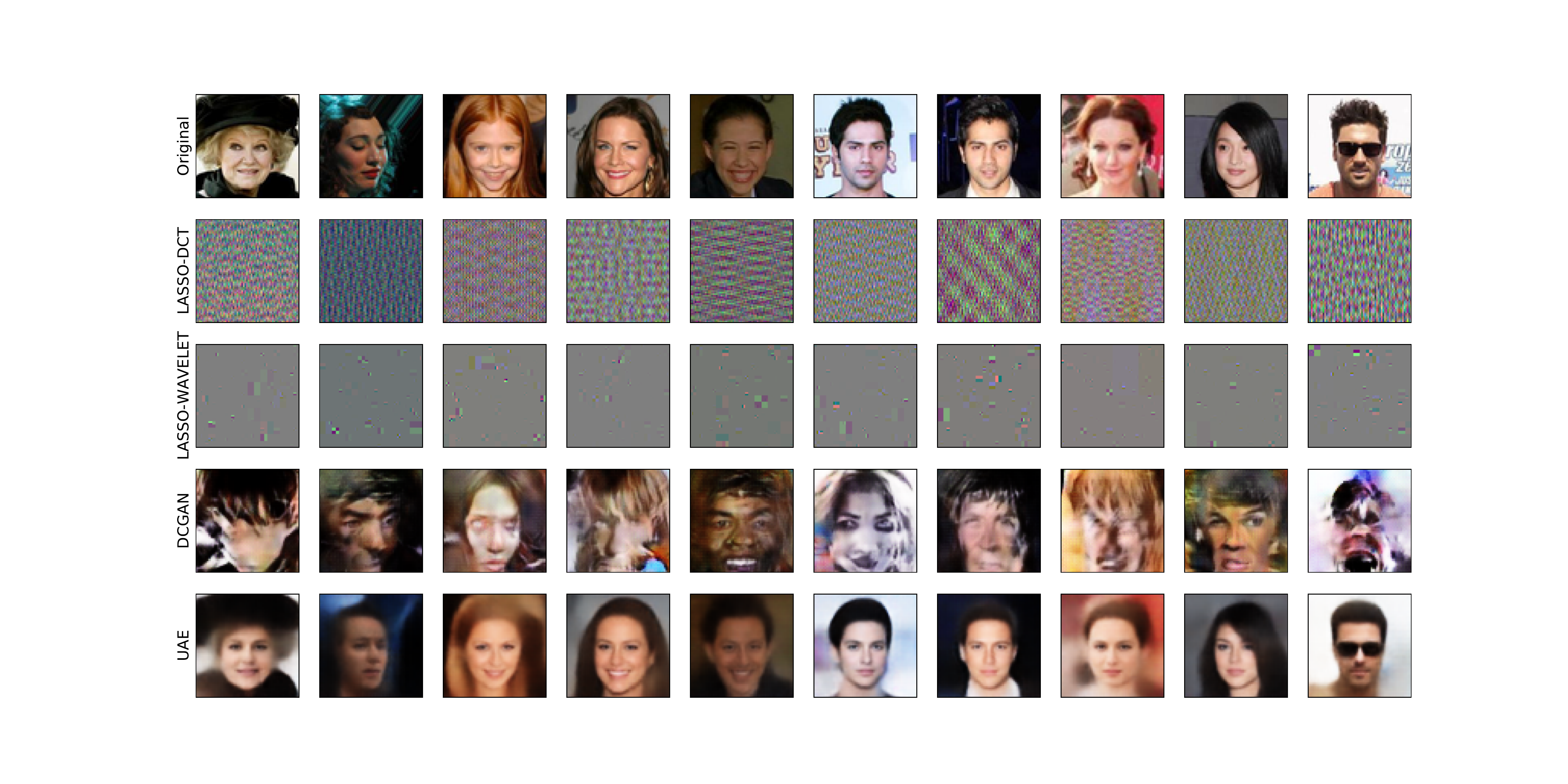}
\caption{Reconstructions for $m=50$ on the CelebA dataset. \textbf{Top:} Target. \textbf{Second:} LASSO-DCT. \textbf{Third:} LASSO-Wavelet. \textbf{Fourth:} DCGAN. \textbf{Last:} UAE. }
\label{fig:reconstr_celeba}
\end{figure*}

For the CelebA dataset, the dimensions of the images are $64 \times 64 \times 3$ and $\sigma=0.01$. The naive pixel basis does not augur well for compressed sensing on such high-dimensional RGB datasets. Following \citet{bora2017compressed}, we experimented with the Discrete Cosine Transform (DCT) and Wavelet basis for the LASSO baseline. Further, we used the DCGAN architecture~\citep{radford2015unsupervised} as in  \citet{bora2017compressed} as our main baseline. For the UAE approach, we used additional convolutional layers in the encoder to learn a $256$ dimensional feature space for the image before projecting it down to $m$ dimensions. 

\textit{Encoder architecture:}
\begin{align*}
\text{Signal}
&\rightarrow\text{Conv[Kernel: 4x4, Stride: 2, Filters: 32, Padding: Same, Activation: Relu]} \\
&\rightarrow \text{Conv[Kernel: 4x4, Stride: 2, Filters: 32, Padding: Same, Activation: Relu]} \\
&\rightarrow \text{Conv[Kernel: 4x4, Stride: 2, Filters: 64, Padding: Same, Activation: Relu]} \\
&\rightarrow \text{Conv[Kernel: 4x4, Stride: 2, Filters: 64, Padding: Same, Activation: Relu]} \\
&\rightarrow \text{Conv[Kernel: 4x4, Stride: 1, Filters: 256, Padding: Valid, Activation: Relu]} \\
&\rightarrow \text{Fully\_Connected[Units: m, Activation: None]}
\end{align*}

\textit{Decoder architecture:}
\begin{align*}
\text{Measurements}
&\rightarrow \text{Fully\_Connected[Units: 256, Activation: Relu]}\\
&\rightarrow \text{Conv\_transpose[Kernel: 4x4, Stride: 1, Filters: 256, Padding: Valid, Activation: Relu]} \\
&\rightarrow \text{Conv\_transpose[Kernel: 4x4, Stride: 2, Filters: 64, Padding: Same, Activation: Relu]} \\
&\rightarrow \text{Conv\_transpose[Kernel: 4x4, Stride: 2, Filters: 64, Padding: Same, Activation: Relu]} \\
&\rightarrow \text{Conv\_transpose[Kernel: 4x4, Stride: 2, Filters: 32, Padding: Same, Activation: Relu]} \\
&\rightarrow \text{Conv\_transpose[Kernel: 4x4, Stride: 2, Filters: 3, Padding: Valid, Activation: Sigmoid]} 
\end{align*}

We consider $m=\{20, 50, 100, 200, 500\}$ measurements. The results are shown in Figure~\ref{fig:celeba}. While the performance of DCGAN is comparable with that of UAE for $m=500$, UAE outperforms DCGAN significantly when $m$ is low. The LASSO baselines do not perform well, consistent with the observations made for the experiments on the MNIST and Omniglot datasets. Qualitative evaluations are shown in Figure~\ref{fig:reconstr_celeba} for $m=50$ measurements.

\newpage
\section{Additional related work}\label{app:related}

\textbf{Dictionary learning.} An uncertainty autoencoder can be also seen as a more flexible, generalized form of \textit{undercomplete} dictionary learning with non-linear encoding and decoding. To see the connection, consider the simplified noise-free setting where the decoding distribution is a Gaussian with fixed variance and the mean of the decoding function is linear in a linear function of the measurements. That is, we are considering a standard \textit{linear} autoencoder with $Y=WX$ and $P_\theta(X \vert Y) = \mathcal{N}(\widehat{W}Y, \Sigma)$, where $\widehat{W}$ is some decoding matrix. Under these assumptions, the UAE objective simplifies to:
\begin{align*}
\min_{W, \widehat{W}}\mathbb{E}_{x \sim Q_{\mathrm{data}}}\left[\Vert x - \widehat{W} W x  \Vert_2^2\right].
\end{align*}
If we think of the decoding $\widehat{W}$ as a dictionary and the encoding $WX$ as the representation then we arrive at an undercomplete dictionary learning. A large body of prior research has focussed on \textit{overcomplete} dictionary learning for compressed sensing. Here, the goal is to learn an \textit{encoding dictionary} and an overcomplete basis in which the original signal is sparse. This basis allows us to leverage algorithms for compressed sensing that are designed based on sparsity assumptions over the signals (see \citep{chen2015compressed} and references therein). An uncertainty autoencoder makes no sparsity assumptions, and it crucially learns a \textit{decoding dictionary} and an encoding basis. By adding more (non-linear) layers to the encoder, one could also learn an (over/under) complete basis for the dataset with desired properties such as sparsity.

\textbf{Further applications of variational information maximization.} The variational information maximization principle underlies many recent algorithms and tasks, such as feature selection~\citep{gao2016variational}, interpretable representation learning in generative adversarial networks~\citep{chen2016infogan,li2017infogail}, and intrinsic motivation in reinforcement learning~\citep{mohamed2015variational}.

\end{document}